%% file: main.tex
\documentclass[10pt,twocolumn,letterpaper]{article}

\usepackage[pagenumbers]{cvpr} 

\definecolor{cvprblue}{rgb}{0.21,0.49,0.74}
\usepackage[pagebackref,breaklinks,colorlinks,allcolors=cvprblue]{hyperref}

\usepackage{amssymb}
\usepackage{amsmath}
\usepackage{bm} 
\usepackage{booktabs}
\usepackage{subcaption}
\usepackage[symbol]{footmisc}
\usepackage{pifont}
\usepackage{multirow}
\usepackage{tabularx}
\definecolor{lightblue}{RGB}{234,249,255}
\usepackage{makecell}

\usepackage{amsfonts}
\usepackage{bm}
\usepackage{comment}
\usepackage{color}
\usepackage{array}

\usepackage{colortab}
\usepackage{colortbl}
\usepackage{arydshln}
\usepackage{multirow}
\usepackage{wrapfig}
 
\usepackage{pifont}
\usepackage{xcolor}

\usepackage[normalem]{ulem}
\useunder{\uline}{\ul}{}

\usepackage{listings}

\newcolumntype{C}{>{\centering\arraybackslash}X}

\newcommand{\mysub}[1]{\textsubscript{#1}}

\title{UniLayDiff: A Unified Diffusion Transformer \\for Content-Aware Layout Generation}

\author{
Zeyang Liu$^{1}$ \quad
Le Wang$^{1}$\thanks{Corresponding author.} \quad
Sanping Zhou$^{1}$ \quad
Yuxuan Wu$^{1}$ \quad \\
Xiaolong Sun$^{1}$ \quad
Gang Hua$^{2}$ \quad
Haoxiang Li$^{3}$
\\[0.5em]
{\small \textsuperscript{1} College of Artificial Intelligence, Xi'an Jiaotong University} \ \ \
{\small \textsuperscript{2} Amazon.com, Inc.}  \ \ \
{\small \textsuperscript{3} Pixocial Technology} \\
}

\begin{document}
\maketitle
\input{sec/0_abstract}    
\input{sec/1_intro}
\input{sec/2_related}

\input{sec/3_method}
\input{sec/4_exp}
\input{sec/5_conclusion}

{
    \small
    \bibliographystyle{ieeenat_fullname}
    \bibliography{main}
}
\input{sec/X_suppl}


\end{document}

%% file: sec/0_abstract.tex
\begin{abstract}
Content-aware layout generation is a critical task in graphic design automation, focused on creating visually appealing arrangements of elements that seamlessly blend with a given background image. The variety of real-world applications makes it highly challenging to develop a single model capable of unifying the diverse range of input-constrained generation sub-tasks, such as those conditioned by element types, sizes, or their relationships. Current methods either address only a subset of these tasks or necessitate separate model parameters for different conditions, failing to offer a truly unified solution. In this paper, we propose UniLayDiff: a Unified Diffusion Transformer, that for the first time, addresses various content-aware layout generation tasks with a single, end-to-end trainable model. Specifically, we treat layout constraints as a distinct modality and employ Multi-Modal Diffusion Transformer framework to capture the complex interplay between the background image, layout elements, and diverse constraints. Moreover, we integrate relation constraints through fine-tuning the model with LoRA after pretraining the model on other tasks. Such a schema not only achieves unified conditional generation but also enhances overall layout quality. Extensive experiments demonstrate that UniLayDiff achieves state-of-the-art performance across from unconditional to various conditional generation tasks and, to the best of our knowledge, is the first model to unify the full range of content-aware layout generation tasks. 
\end{abstract}

%% file: sec/1_intro.tex
\section{Introduction}



Layout generation in graphic design and UI automation arranges visual elements (e.g., text, logos) to create aesthetically pleasing advertisements, posters~\cite{guo2021vinci, hsu2023pku}, and magazines~\cite{jahanian2013recommendation, yang2016automatic}. This task's complexity arises from diverse creative requirements, including unconditional generation, and generation conditioned on element types, sizes, inter-element relations (e.g., ``text A is larger than logo B"), layout refinement, and partial layout completion.

\begin{figure}[t]
\begin{center}
\includegraphics[width=\linewidth]{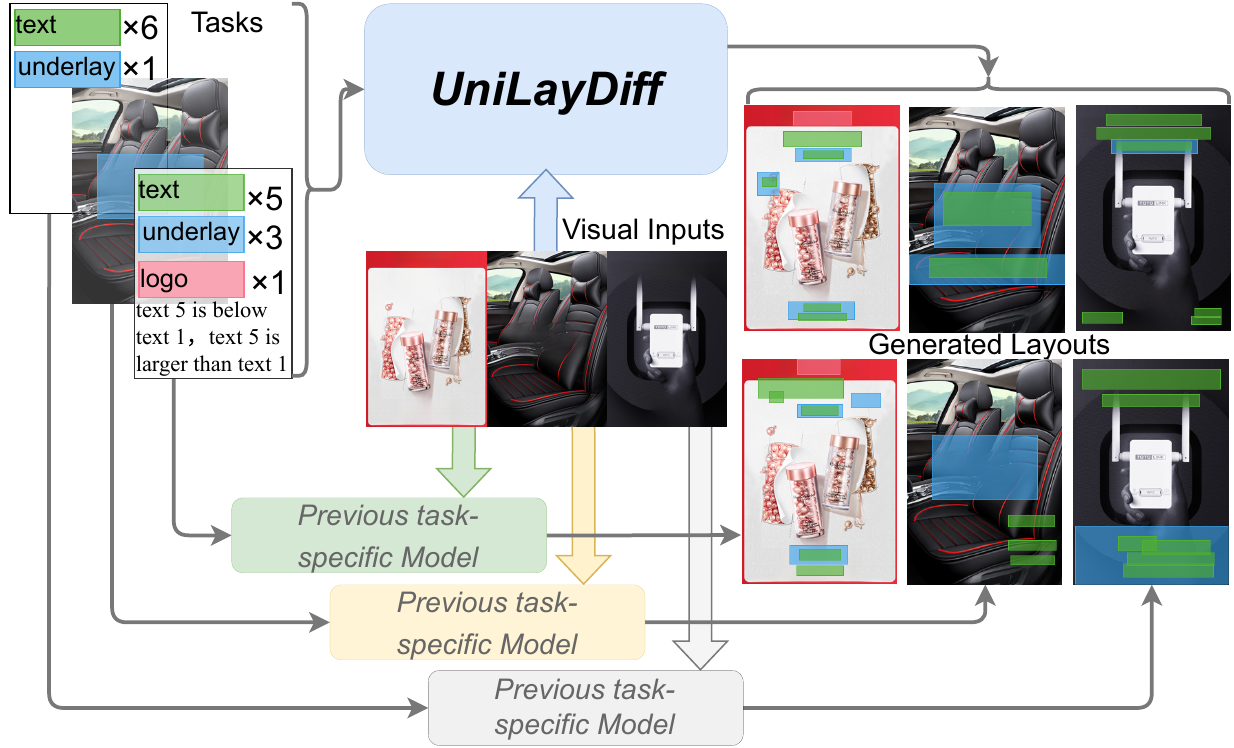}
\end{center}
\vspace{-0.5cm}

   \caption{Previous models like RALF and LayoutDiT handle different conditional generation tasks with different parameters or only support a part of conditional tasks, whereas our proposed model, UniLayDiff, is able to handle various content-aware conditional generation tasks in a unified diffusion model.}
\vspace{-1.25\baselineskip}

\label{fig:intro}
\end{figure}




Generally, layout generation can be categorized into two types: content-agnostic and content-aware. 
The content-agnostic task generates layouts without considering the background image or its content, while content-aware layout generation aims to generate a fitting layout that harmonizes with the given background image.
In recent years, various generative models have made notable progress, including GAN~\cite{li2019layoutgan, li2020attribute, hsu2023posterlayout, xu2023unsupervised}, VAE~\cite{jyothi2019layoutvae, yamaguchi2021canvasvae, cao2022geometry}, and Transformer-based models~\cite{gupta2021layouttransformer, horita2023retrieval} in these two scenarios.
More recent studies, however, have shifted the focus towards the application of diffusion models for better generative quality and versatility in conditional generation.
Hence, they have been adopted to develop unified models that adapt to various conditions and improve generation quality, especially in the content-agnostic setting~\cite{inoue2023layoutdm, hui2023unifying, chen2024towards, guerreiro2025layoutflow}.
These diffusion-based unified models have shown better consistency and generalization ability across different tasks.

Nevertheless, unification in the content-aware scenario remains inadequately explored.
When extending unified models from the content-agnostic domain to the content-aware setting, they fail due to intricate relationships between layouts, various types of conditions, and the background canvas.
Although LayoutDiT~\cite{li2024cgbdmcontentgraphicbalance} achieves content-aware generation with various conditions in the same model, it requires retraining the model with different parameters for different condition specifications, as shown in Fig.~\ref{fig:intro}, limiting its generalization ability and parameter efficiency.
These limitations motivate the development of a truly unified approach for content-aware layout generation.

To tackle the aforementioned challenges and achieve unification in the content-aware setting, we propose UniLayDiff, a novel framework built upon the powerful diffusion paradigm, as shown in Fig.~\ref{fig:intro} and Fig.~\ref{fig:model}.
Specifically, we treat layout constraints as a distinct input modality alongside visual information from the background image, and integrate a Multi-Modal Attention (MM-Attention) mechanism~\cite{esser2024mmdit}, inspired by recent successes in large-scale image generation~\cite{li2024playground,zhang2024creatilayout,Li_2025_CVPR,Wang_2025_CVPR}, into our transformer backbone. 
This design enables our model to flexibly learn the high-dimensional relationships between image content, explicit user constraints, and final layout structure within a single unified architecture.
Moreover, incorporating relational constraints is particularly challenging, as joint training can degrade performance on other tasks, as shown in Tab.~\ref{tab:relationshipinjection} and Fig.~\ref{fig:abltune}. To overcome this, we introduce relational conditioning via Low-Rank Adaptation (LoRA)~\cite{hu2022lora} fine-tuning. After pre-training on all tasks except relations, we apply LoRA on the Partial Constraint Mask branch in our model to extend the robust generative prior with relation-specific adaptations. Such a paradigm preserves overall quality while endowing the model with strong performance across diverse conditional generation tasks.




We evaluate our method on two public benchmarks PKU~\cite{hsu2023pku} and CGL~\cite{zhou2022cgl}. Extensive experiments demonstrate that our model achieves state-of-the-art performance. The Multi-Modal Transformer can effectively capture the complex relationships between background image and layout constraints. LoRA fine-tuning effectively integrates relation constraints in content-aware generation.

Overall, the contributions of our work are the following:

\begin{itemize}
\item We propose UniLayDiff, a unified framework based on a Multi-Modal Diffusion Transformer that uses a single model to address the majority of content-aware layout generation tasks, from unconditional generation to refinement and completion.
\item In order to avoid performance degradation when directly joint-training with relation conditions, we introduce relation constraints through LoRA fine-tuning, which effectively extends the robust generative prior to achieve relation‑conditioned layout generation.
\item To the best of our knowledge, UniLayDiff is the first truly unified model in content-aware layout generation that can handle the full spectrum of input-constrained sub-tasks using a single set of weights, achieving state-of-the-art results on multiple benchmarks.

\end{itemize}

%% file: sec/2_related.tex
\section{Related Work}
\subsection{Content-agnostic Layout Generation}
Content-agnostic layout generation~\cite{shi2023intelligent} arranges elements based on abstract rules like category or size, without a background image. The field has evolved from early task-specific generative models like GANs~\cite{li2019layoutgan, li2020attribute}, VAEs~\cite{jyothi2019layoutvae, yamaguchi2021canvasvae}, and autoregressive Transformers~\cite{gupta2021layouttransformer} toward more versatile, unified frameworks.
The pursuit of unification led to various models, including autoregressive approaches like LayoutFormer++~\cite{jiang2023layoutformer++} and discrete diffusion models~\cite{hui2023unifying, inoue2023layoutdm}. However, these can be limited by sequential generation or suboptimal conditioning methods that result in poor constraint adherence. While more recent methods using continuous diffusion or flow matching, such as LACE~\cite{chen2024towards} and LayoutFlow~\cite{guerreiro2025layoutflow}, offer advantages like integrating differentiable aesthetic losses, they critically lack native support for complex relational constraints~\cite{chen2024towards, guerreiro2025layoutflow}. This often necessitates a separate post-processing pipeline~\cite{chen2024towards}, revealing a crucial gap in robustly modeling inter-element relations for sophisticated layouts.

\subsection{Content-aware Layout Generation}
Content-aware layout generation, which arranges elements in harmony with a background image, is a complex multi-modal task.  ContentGAN~\cite{zheng19tog} was the first to tackle this by incorporating the image semantics of the input canvas. Subsequently, CGL-GAN~\cite{zhou2022cgl} used a GAN conditioned on image saliency, DS-GAN~\cite{hsu2023pku} employed a CNN-LSTM framework, and ICVT~\cite{icvt} utilized a conditional VAE.  RADM~\cite{li2023relation} and ContentDM~\cite{guo2025contentdm} leveraged diffusion models. Scan-and-Print~\cite{Hsu-IJCAI2025-ScanandPrint} is a lightweight autoregressive model using patch-level ``Scan" for efficient data summarization and ``Print" for data augmentation.  However, these approaches lack support for various subtasks. While some models like retrieval-augmented RALF~\cite{horita2023retrieval} and diffusion-based LayoutDiT~\cite{li2024cgbdmcontentgraphicbalance} attempt multiple tasks, they remain task-specific, requiring separate parameters and falling short of unification. LayoutDiT also lacks support for relation-constrained generation.
Recently, LLM-based models~\cite{chen2025posta,seol-2024-ECCV-posterllama,lin-2023-NIPS-layoutprompter,Hsu-CVPR2025-postero} have also emerged. These methods demonstrate promising capabilities in layout generation, though they suffer from high inference costs.

%% file: sec/3_method.tex
\section{Methodology}

\begin{figure*}[t]
    \centering 
    \includegraphics[width=1.0\linewidth]{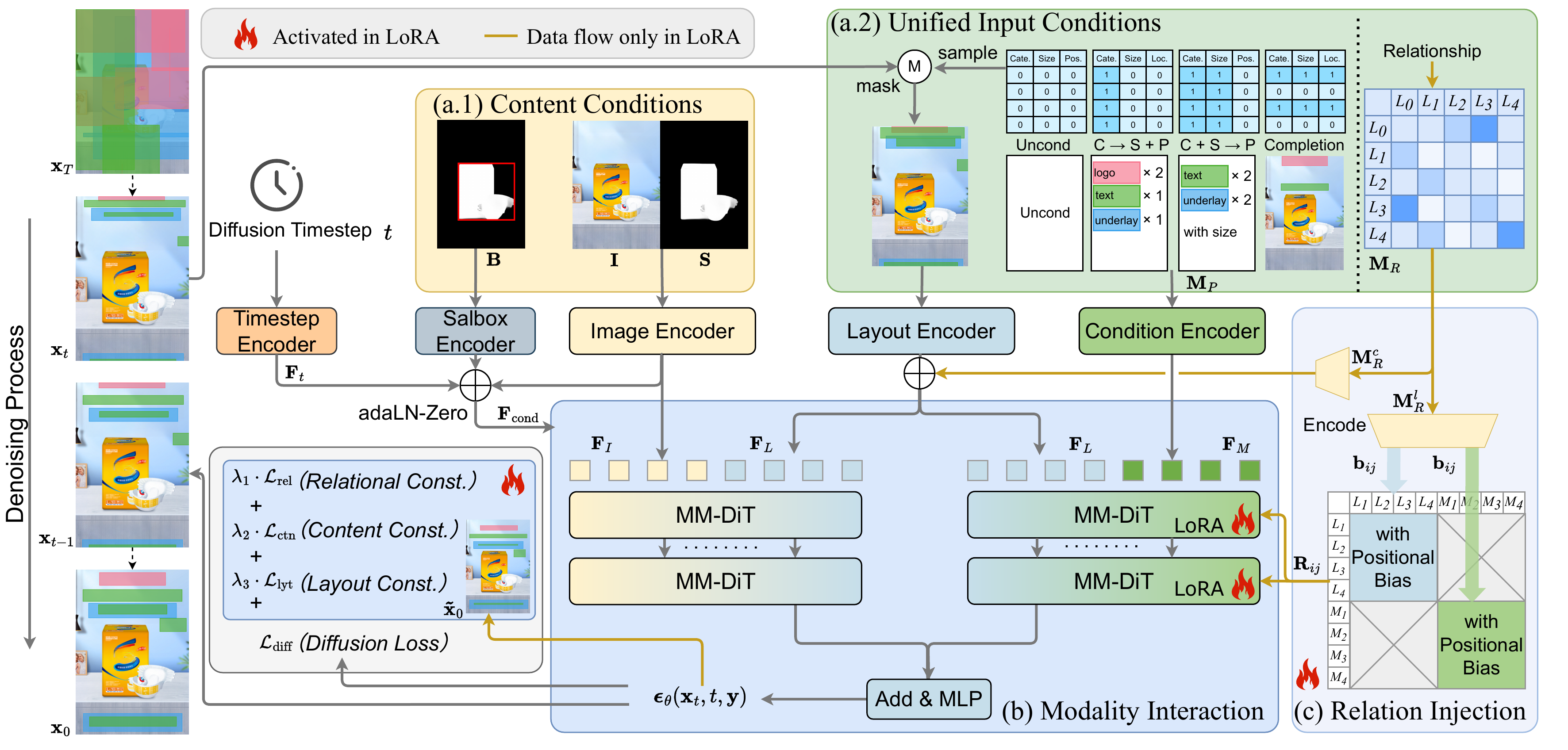}
    \vspace{-0.5cm}
    \caption{%
Overview of UniLayDiff. (a) Our framework takes inputs including the image \(\mathbf{I}\) with its saliency map \(\mathbf{S}\), a noisy layout \(\mathbf{x}_t\), a partial constraint mask \(\mathbf{M}_P\) (sampled from various tasks), and a relation matrix \(\mathbf{M}_R\). (b)  Features \(\mathbf{F}_I\), \(\mathbf{F}_L\), and \(\mathbf{F}_M\) interact via the dual-path MM-Attention module. (c) Relational constraints are integrated via a block-diagonal relative positional bias (where \({L}_i\) denotes the $i$-th layout element,\({M}_i\) the $i$-th mask element, and \({L}_0\) represents the canvas). This relation module, and parts of the mask interaction branch, are activated during the LoRA fine-tuning stage. Our model is first pre-trained on four fundamental tasks (a.2), then fine-tuned for the relation task and enhanced quality. The resulting, fine-tuned model is used for inference on all tasks.
}

    \vspace{-0.5cm}
    \label{fig:model}
\end{figure*}

Our goal is to develop a unified content-aware layout generation framework, \textbf{UniLayDiff}, able to handle a full spectrum of conditional generation tasks with a single model, as illustrated in Fig.~\ref{fig:model}.

\subsection{Preliminary: Diffusion Models}

Denoising Diffusion Probabilistic Models (DDPMs)~\cite{ho2020denoising} are generative models that learn to reverse a forward Markov process that gradually adds Gaussian noise to data. The forward process is defined by the transition $q(\mathbf{x}_t | \mathbf{x}_{t-1}) = \mathcal{N}(\mathbf{x}_t; \sqrt{1 - \beta_t}\mathbf{x}_{t-1}, \beta_t\mathbf{I})$, where $\{\beta_t\}_{t=1}^T$ is a predefined variance schedule. This allows sampling a noisy version $\mathbf{x}_t$ of any clean data $\mathbf{x}_0$ at an arbitrary timestep $t$ in a closed form:
\begin{equation}
    \mathbf{x}_t = \sqrt{\bar{\alpha}_t}\mathbf{x}_0 + \sqrt{1 - \bar{\alpha}_t}\bm{\epsilon},
\label{eq:sample}
\end{equation}
where $\alpha_t = 1 - \beta_t$, $\bar{\alpha}_t = \prod_{s=1}^t \alpha_s$, and $\bm{\epsilon} \sim \mathcal{N}(\mathbf{0}, \mathbf{I})$ is the sampled noise. The reverse process employs a neural network, $\bm{\epsilon}_\theta(\mathbf{x}_t, t, \mathbf{y})$, trained to predict the original noise $\bm{\epsilon}$ given the noisy data $\mathbf{x}_t$, the timestep $t$, and an optional condition $\mathbf{y}$. The model is optimized by minimizing a simplified L2 loss on the predicted noise:
\begin{equation}
    \mathcal{L}_{\text{diff}} = \mathbb{E}_{t, \mathbf{x}_0, \mathbf{y}, \bm{\epsilon}} \left[ || \bm{\epsilon} - \bm{\epsilon}_\theta(\mathbf{x}_t, t, \mathbf{y}) ||^2 \right].
\label{diffusion_loss}
\end{equation}


\subsection{Unified Conditioning}
To build a unified content-aware layout generation model, we formulate all the generation sub-tasks as a conditional denoising problem. 

\noindent \textbf{Content conditions.}
In the content-aware setting, the generation process is guided by visual information. Following prior works \cite{horita2023retrieval,li2024cgbdmcontentgraphicbalance}, our model takes two visual input: the RGB image of the canvas, $\mathbf{I} \in \mathbb{R}^{H\times W\times 3}$, and its corresponding saliency map, $\mathbf{S} \in \mathbb{R}^{H\times W\times 1}$, where $H$ and $W$ are the height and width of the canvas. The saliency map is obtained using an off-the-shelf saliency detection method~\cite{qin2019basnet,qin2022highly}. Furthermore, following prior work~\cite{li2024cgbdmcontentgraphicbalance}, we extract bounding boxes $\mathbf{B}$ from the saliency map and feed them into the model as an additional guidance. 

\noindent \textbf{Unified input conditions.}
 We summarize the diverse conditions into two primary types: \textit{Partial Layout Constraints} and \textit{Relational Constraints}. Partial constraints, used for tasks like completion and generation conditioned on element types, are represented by a Partial Constraint Mask $\mathbf{M}_{P} \in \{0,1\}^{N \times (C + 4)}$, where $N$ is the number of layout elements, $C$ is the number of element categories. Relational constraints, which define both positional (e.g., `left of') and size (e.g., `larger than') relationships among elements and the canvas, are encoded in a relation matrix $\mathbf{M}_{R} \in \mathbb{Z}^{(N + 1) \times (N + 1) \times 2}$.

 Critically, unlike prior task-specific methods~\cite{horita2023retrieval,li2024cgbdmcontentgraphicbalance} that directly use partially ``clean" layouts as model input, which limits their unification capabilities, we treat both $\mathbf{M}_{P}$ and $\mathbf{M}_{R}$ as explicit, additional conditions. These are fed directly into our network's attention mechanism, empowering our model to flexibly support a full spectrum of tasks within a single, coherent framework. Therefore, in our unified model, the condition $\mathbf{y}$ in Eq.~\ref{diffusion_loss} can be represented as: $\mathbf{y} = \{\mathbf{I},\,\mathbf{S},\,\mathbf{B},\,\mathbf{M}_{R},\,\mathbf{M}_{P}\}$.

\subsection{Model Architecture}\label{sec:model_arc}
To implement our unified modeling approach, we propose a unified architecture, based on Multi-Modal Diffusion Transformer (MM-DiT)~\cite{esser2024mmdit}. Modality-specific encoders project all conditions into a common embedding space. The model then uses an MM-Attention mechanism to handle image-layout and mask-layout interactions, enabling flexible task handling. 

\noindent \textbf{Modality encoding.}
To condition the denoising process, we first project all input modalities into feature embeddings. This is accomplished through a set of specialized encoders. Specifically, following~\cite{li2024cgbdmcontentgraphicbalance}, a ViT~\cite{dosovitskiy2020image}-based image encoder takes the concatenated image $\mathbf{I}$ and saliency map $\mathbf{S}$ to extract visual features $\mathbf{F}_{I}$. The layout $\mathbf{L}$ and Partial Constraint Mask $\mathbf{M}_{P}$ are processed by separate encoders that first project their geometric and categorical components independently, then sum the resulting embeddings to produce layout features $\mathbf{F}_{L}$ and mask features $\mathbf{F}_{M}$. 
The visual features $\mathbf{F}_{I}$, layout features $\mathbf{F}_{L}$, and Partial Constraint Mask features $\mathbf{F}_{M}$ are later fused within the MM-DiT.

\noindent \textbf{Modality interaction.}
To better guide the generation, we elevate the Partial Constraint Mask $\mathbf{M}_{P}$ to a distinct modality, treating it as a structural skeleton. Our approach contrasts with previous content-agnostic works~\cite{chen2024towards,guerreiro2025layoutflow} where the mask was merely part of the layout's initial encoding. To fuse the image, layout, and this new mask modality, we employ a MM-DiT-based architecture. (See Fig.~\ref{fig:model}(b).)

The MM-DiT leverages a MM-Attention mechanism to model interactions across modalities. For a set of $M$ modality token embeddings $\{\mathbf{f}^{(i)}\}_{i=1}^M$, the combined query, key, and value matrices are constructed by concatenating the linearly projected tokens from each modality:
\begin{equation}
\mathrm{Attention}(\mathbf{Q},\mathbf{K},\mathbf{V}) = \mathrm{softmax}\left(\frac{\mathbf{Q}\mathbf{K}^\top}{\sqrt{d}}\right)\mathbf{V},
\end{equation}
\begin{align}
    \mathbf{Q} &= [\mathbf{P}_Q^{(1)}\mathbf{f}^{(1)}; \dots; \mathbf{P}_Q^{(N)}\mathbf{f}^{(M)}], \notag \\
    \mathbf{K} &= [\mathbf{P}_K^{(1)}\mathbf{f}^{(1)}; \dots; \mathbf{P}_K^{(N)}\mathbf{f}^{(M)}],  \\
    \mathbf{V} &= [\mathbf{P}_V^{(1)}\mathbf{f}^{(1)}; \dots; \mathbf{P}_V^{(N)}\mathbf{f}^{(M)}], \notag
\end{align}
where $[\cdot;\cdot]$ denotes concatenation along the token dimension,  $\mathbf{P}^{(i)}$ are linear projections.

However, directly fusing three modalities (image, layout, and mask) in this manner resulted in suboptimal performance, as demonstrated in our ablation study (Tab.~\ref{tab:unifiedtrainingmmdit}). This is likely due to the optimization challenges when processing three disparate modalities within a single MM-Attention mechanism, which may dilute key interactions~\cite{zhang2024creatilayout}. 
To address this, we adopt two parallel MM‑DiT branches: one captures interactions between visual features (\(\mathbf{F}_I\)) and layout features (\(\mathbf{F}_L\)), and the other between mask features (\(\mathbf{F}_M\)) and layout features (\(\mathbf{F}_L\)). The final layout representation is obtained by element‑wise summing the layout-token outputs from both branches.


Moreover, a conditioning vector $\mathbf{F}_{\text{cond}}$ is created by summing the time embedding $\mathbf{F}_{t}$, embedding of bounding box $\mathbf{B}$ and a global visual feature obtained by linearly projecting $\mathbf{F}_{I}$ across its token dimension. This vector guides the denoising network by being integrated into both parallel branches 
through adaptive layer norm - Zero (adaLN-Zero) used in DiT~\cite{peebles2023scalable}.

\subsection{Imposing Relation Constraints}

As a unified model, it also needs to be capable of generating layouts under relational constraints. To achieve this, we inject relations via relative positional bias and introduce an additional relational constraint for explicit enforcement.

\noindent \textbf{Relation injection.}
To incorporate relational constraints, we split the relation matrix $\mathbf{M}_{R}$ precomputed from the ground-truth layout, into layout-canvas relations ($\mathbf{M}_{R}^c$) and intra-layout relations ($\mathbf{M}_{R}^{l}$). The layout-canvas relations $\mathbf{M}_R^c$ are encoded via a learnable embedding and added element-wise to the layout token embeddings ($\mathbf{F}_L$). For intra-layout relations, as shown in Fig.~\ref{fig:model}(c), we introduce a masked relative positional bias mechanism to inject them into MM-Attention, inspired by relative positional bias\cite{2020t5}.



To embed spatial priors, we inject a block-diagonal relative positional bias $\mathbf{R}$, into the MM-Attention mechanism, confining its influence to intra-modal relationships. The attention logits are computed as:
\begin{equation}
\text{logit}_{ij} = \frac{\mathbf{q}_i \mathbf{k}_j^T}{\sqrt{d}} + \mathbf{R}_{ij}.
\end{equation}
The bias $\mathbf{R}_{ij}$ is non-zero only when tokens $i$ and $j$ belong to the same modality, in which case it is derived from learnable embeddings of their pairwise geometric relations:
\begin{equation}
    \mathbf{R}_{ij} =
    \begin{cases}
        \mathbf{b}_{ij}, & \text{if } \text{modality}(i) = \text{modality}(j) \\
        0, & \text{otherwise},
    \end{cases}
\label{eq:attnmask}
\end{equation}
where the modality-specific bias is computed as $\mathbf{b}_{ij} = \text{Emb}_{\text{size}}(\Delta_{ij}^{\text{size}}) + \text{Emb}_{\text{pos}}(\Delta_{ij}^{\text{pos}})$, $\Delta^{\text{size}}$ and $\Delta^{\text{pos}}$ denote the relative size and position parts of $\mathbf{M}_{R}^{l}$, $\text{Emb}_{\text{size}}$ and $\text{Emb}_{\text{pos}}$ are learnable embedding layers. This targeted structural bias allows the model to infer spatial arrangements within a given modality without disrupting the learned cross-modal attention patterns.



\noindent \textbf{Relational constraint.} 
As shown in the ablation study (Tab.~\ref{tab:constraints}), we found that relying solely on relative position encoding is insufficient for enforcing complex relationships, particularly for size constraints. 
Hence, we introduce a relational loss, $\mathcal{L}_{\mathrm{rel}}(\tilde{\mathbf{x}}_0, \mathbf{x}_0)$. This loss imposes explicit supervision by translating discrete geometric constraints from $\mathbf{x}_0$ into a differentiable penalty, which is applied to the prediction $\tilde{\mathbf{x}}_0$ at each time step to penalize violations.

Specifically, the closed-form sampling formulation for the forward process denoted in Eq.~\ref{eq:sample} yields a way to estimate $x_0$ at each time step
using the predicted noise ${\epsilon}_{\theta}$ as:
\begin{equation}
\tilde{\mathbf{x}}_0(\mathbf{x}_t) = ({\mathbf{x}_t - \sqrt{1 - \bar{\alpha}_t} \cdot \boldsymbol{\epsilon}_\theta(\mathbf{x}_t, t, \mathbf{y})}) / {\sqrt{\bar{\alpha}_t}}.
\label{eq:x0_pred}
\end{equation}
The relational loss $\mathcal{L}_{\mathrm{rel}}(\tilde{\mathbf{x}}_0, \mathbf{x}_0)$ operates by comparing properties extracted from 
the $\tilde{\mathbf{x}}_0$ estimated from $\boldsymbol{\epsilon}_\theta(\mathbf{x}_t, t, \mathbf{y})$ 
against ground-truth relations derived from the original data $\mathbf{x}_0$.
Specifically, a set of predicted bounding boxes, $\{ \tilde{\mathbf{b}}_1, \dots, \tilde{\mathbf{b}}_N \}$, is parsed from $\tilde{\mathbf{x}}_0$. For any pair of these boxes $(\tilde{\mathbf{b}}_i, \tilde{\mathbf{b}}_j)$, we compute the log-ratio of their areas, $d_{ij} = \log A(\tilde{\mathbf{b}}_j) - \log A(\tilde{\mathbf{b}}_i)$, where $A(\cdot)$ denotes the function that computes the area of a bounding box. This formulation is scale-invariant, focusing on relative size differences rather than absolute pixel values. Based on this ratio, we formulate logits for three size relationships (`smaller', `equal', `larger') using a tolerance margin $\alpha$:
\begin{equation}
\resizebox{\columnwidth}{!}{$
\begin{split}
    \mathbf{z}_{ij}^{(\text{size})} = \bigg[ & \frac{\log(1-\alpha) - d_{ij}}{\tau_{rel}},  \frac{\log(1+\alpha) - |d_{ij}|}{\tau_{rel}},
    \frac{d_{ij} - \log(1+\alpha)}{\tau_{rel}} \bigg],
\end{split}
$}
\end{equation}
where $\tau_{rel}$ is a temperature parameter. The final loss is the masked cross-entropy between these predicted logits $\mathbf{z}_{ij}^{(\text{size})}$ and the ground-truth labels $\mathbf{y}_{ij}^{(\text{size})}$, which are themselves derived from the ground-truth data $\mathbf{x}_0$.

\subsection{LoRA Fine-tuning} \label{sec:LoRA}


We observed that training the relation task from scratch degrades performance, likely due to a conflict between its discrete constraints and the continuous diffusion process. (More details are shown in Tab.~\ref{tab:relationshipinjection} and its accompanying analysis.) Therefore, we adopt a pre-training and LoRA fine‑tuning strategy, enabling the model to learn relational constraints while retaining its strong generative prior.

Our model is pre-trained on four fundamental tasks to learn general layout generation inspired by recent content-agnostic work~\cite{chen2024towards,guerreiro2025layoutflow}, 
as shown in Fig.~\ref{fig:model}(a.2). 
During pre‑training, we randomly sample examples from these tasks in each batch. We only adopt $\mathcal{L}_{\mathrm{diff}}$ as the objective function for this stage.

Diverging from typical plug-and-play adaptation~\cite{incontextlora,zhang2025easycontrol}, we employ LoRA as a fine-tuning stage of our unified training process. In this stage, we freeze most pre-trained weights, activating only our relation injection module for training. Simultaneously, we update the attention linear projections of the condition or Partial Constraint Mask branch
using Low-Rank Adaptation (LoRA)~\cite{hu2022lora}. This method keeps the original weights $\mathbf{P}$ frozen and introduces trainable low-rank matrices, formulated as $\mathbf{P}' = \mathbf{P} + \alpha \cdot \mathbf{B}\mathbf{A}$, where $r \ll \min(d, k)$. Finally, the relation task is added to the sampling pool to be learned alongside the original pre-training tasks.

Moreover, fine-tuning via LoRA enables us to impose additional constraints to further improve generation quality. Specifically for content-aware settings, it's vital to preserve the salient regions of the background. To encourage layouts not obscuring important visual areas, we design a content-constraint loss, $\mathcal{L}_{\mathrm{ctn}}(\tilde{\mathbf{x}}_0, \mathbf{S})$. The effectiveness of this loss in reducing occlusion is validated in Tab.~\ref{tab:constraints}.
This function uses 
$\tilde{\mathbf{x}}_0$ estimated by $\boldsymbol{\epsilon}_\theta(\mathbf{x}_t, t, \mathbf{y})$ in Eq.~\ref{eq:x0_pred} 
and a given saliency map $\mathbf{S}$, where high values indicate regions to be avoided. 
The loss operates on the set of predicted bounding boxes $\{\tilde{\mathbf{b}}_i\}$ extracted from $\tilde{\mathbf{x}}_0$. For each box $\tilde{\mathbf{b}}_i=(x_1, y_1, x_2, y_2)$, we compute a differentiable sigmoid ``soft" mask, $\mathbf{M}_i$, as a smooth approximation of the hard bounding box to ensure stable gradient flow. The mask's value at each pixel $(u,v)$ is calculated as:
\begin{equation}
\begin{split}
\mathbf{M}_i(u, v) = \sigma( \tau_{sig}(u - x_1)) \cdot \sigma(\tau_{sig}(x_2 - u)) 
\cdot \\\sigma(\tau_{sig}(v - y_1)) \cdot \sigma(\tau_{sig}(y_2 - v)),
\end{split}
\end{equation}
where $\tau_{sig}$ is a temperature parameter, $\sigma(\cdot)$ is the Sigmoid function. The final content-constraint loss is the sum of the average saliency intensities, $\bar{S}_i$, within each soft-masked region over all $N$ predicted boxes:
\begin{equation}
\mathcal{L}_{\mathrm{ctn}} = \sum_{i=1}^{N} \bar{S}_i, \text{where} \bar{S}_i = \frac{\sum_{u,v} \mathbf{S}(u,v) \cdot \mathbf{M}_i(u,v)}{\sum_{u,v} \mathbf{M}_i(u,v) + \delta_{eps}}.
\end{equation}
Here, $\delta_{eps}$ is a small constant added for numerical stability to prevent division by zero.




Overall, the comprehensive loss for LoRA fine-tuning is $\mathcal{L}_{\mathrm{diff}}$ plus weighted sum of the auxiliary losses as following:
\begin{equation}
\mathcal{L}_{\mathrm{LoRA
}} = \mathcal{L}_{\mathrm{diff}} + \lambda_{\mathrm{1}}\cdot\mathcal{L}_{\mathrm{rel}} + \lambda_{\mathrm{2}}\cdot\mathcal{L}_{\mathrm{ctn}} +
\lambda_{\mathrm{3}}\cdot\mathcal{L}_{\mathrm{lyt}} ,
\label{eq:loss_lora}
\end{equation}
where $\mathcal{L}_{\mathrm{lyt}}$ is the layout aesthetic constraint similar with that used by LACE ~\cite{chen2024towards}. $\lambda_{1}$, $\lambda_{2}$, and $\lambda_{3}$ are hyperparameters. Details are in the supplementary material.

%% file: sec/4_exp.tex
\section{Experiments}

\input{table/uncond_gen}

\input{table/cond_gen}


\subsection{Experimental Setup}

\noindent \textbf{Datasets.}
We conduct experiments on two publicly available  e-commerce poster datasets: \textbf{PKU}~\cite{hsu2023pku} and \textbf{CGL}~\cite{zhou2022cgl}, following~\cite{horita2023retrieval,li2024cgbdmcontentgraphicbalance,Hsu-CVPR2025-postero,Hsu-IJCAI2025-ScanandPrint}. The PKU dataset contains three types of layout elements: logo, text, and underlay, while the CGL dataset includes an additional ``embellishment'' category. 
To ensure a fair comparison and enable evaluation on constrained generation tasks, we follow the experimental framework and data split protocol of 
RALF~\cite{horita2023retrieval}. 
For the PKU dataset, we use 7,735 samples for training, 1,000 for validation, and 1,000 for testing. For the CGL dataset, the splits are 48,544, 6,002, and 6,002 for training, validation, and testing, respectively.

\noindent \textbf{Evaluation metrics.}
We adopt metrics from two aspects as in previous studies~\cite{horita2023retrieval, li2024cgbdmcontentgraphicbalance} to comprehensively assess our proposed model.
\begin{itemize}

\item \textbf{Content Metrics.} We measure content harmony using \emph{Occlusion} ($\mathrm{Occ} \downarrow$), which calculates the proportion of layout elements covering salient regions of the canvas, and \emph{Unreadability} ($\mathrm{Rea} \downarrow$), which assesses the visual clutter behind text elements by computing the average spatial gradient of the occupied image area.

 \item \textbf{Graphic Metrics.} Graphic quality is assessed by \emph{Overlay} ($\mathrm{Ove} \downarrow$), the average IoU between non-underlay elements; \emph{Loose} ($\mathrm{Und}\mysub{L} \uparrow$) and \emph{Strict} ($\mathrm{Und}\mysub{S} \uparrow$) \emph{Underlay Effectiveness}, measuring underlay coverage; and \emph{Fr\'echet Inception Distance} ($\mathrm{FID} \downarrow$), the distributional similarity between generated and real layouts~\cite{kikuchi2021constrained}, which we compute using the feature extractor provided by~\cite{horita2023retrieval}. For relational tasks, we also report the \emph{Relation Violation Rate} ($\mathrm{Vio} \downarrow$) to evaluate constraint adherence.

  \end{itemize}

\noindent \textbf{Generation tasks.}
We evaluate our model on a comprehensive set of layout generation tasks to demonstrate its versatility. These tasks include:

\noindent$Uncond$ generates layouts without input constraints.

\noindent$Category \rightarrow Size + Position (C \rightarrow S + P)$ takes categories as input and generates the corresponding elements.

\noindent$Category + Size \rightarrow Position (C + S \rightarrow   P)$ generates element positions based on given element categories and sizes.

\noindent$Completion$ generates a full layout when a small subset of elements are already known and placed.

\noindent$Refinement$ corrects noisy layouts perturbed by Gaussian noise ($\mu=0,\ \sigma=0.01$)~\cite{xu2023unsupervised}. Continuous diffusion models supporting unconditional generation can be directly used for refinement via slight DDIM parameter adjustments~\cite{chen2024towards,li2024cgbdmcontentgraphicbalance}.

\noindent$Relationship$ generates a layout conditioned on both element types and a set of specified spatial relationships between element pairs. 
Following~\cite{horita2023retrieval},
we randomly sample 10\% of the relationships in our experiments.

\noindent \textbf{Comparisons.}
We compare UniLayDiff with the following open‑source methods: CGL-GAN~\cite{zhou2022cgl}, DS-GAN~\cite{hsu2023posterlayout}, ICVT~\cite{icvt}, LayoutDM~\cite{inoue2023layoutdm}\footnote[2]{LayoutDM is originally designed for content-agnostic settings, extended by~\cite{horita2023retrieval} as a baseline.}, RALF~\cite{horita2023retrieval}, LayoutDiT~\cite{li2024cgbdmcontentgraphicbalance}, and Scan-and-Print~\cite{Hsu-IJCAI2025-ScanandPrint}, as well as the LLM-based methods PosterLlama~\cite{seol-2024-ECCV-posterllama}, LayoutPrompter~\cite{lin-2023-NIPS-layoutprompter}, and PosterO~\cite{Hsu-CVPR2025-postero}. For fairness, we adopt results from the reproduction of LayoutDiT, PosterO and Scan-and-Print, re-running the baselines  to obtain missing FID scores and relation task results. Implementation details for UniLayDiff are in the supplementary material.



\subsection{Quantitative Results}



We evaluate the performance of our proposed UniLayDiff against state-of-the-art (SOTA) methods across the six subtasks, demonstrating its superior performance. Tab.~\ref{tab:res_uncond} and Tab.~\ref{tab:res_cond} showcase our model's effectiveness across a wide spectrum of subtasks. Across both PKU and CGL datasets, UniLayDiff sets a new state-of-the-art, outperforming existing methods on the majority of content and graphical metrics. These results validate the effectiveness of our unified framework. More quantitative results are provided in the supplementary material, including Generation performance on the unannotated test set, Cross-dataset evaluation and Computational cost.



\begin{figure*}[t]
    \centering 
    \includegraphics[width=1.0\linewidth]{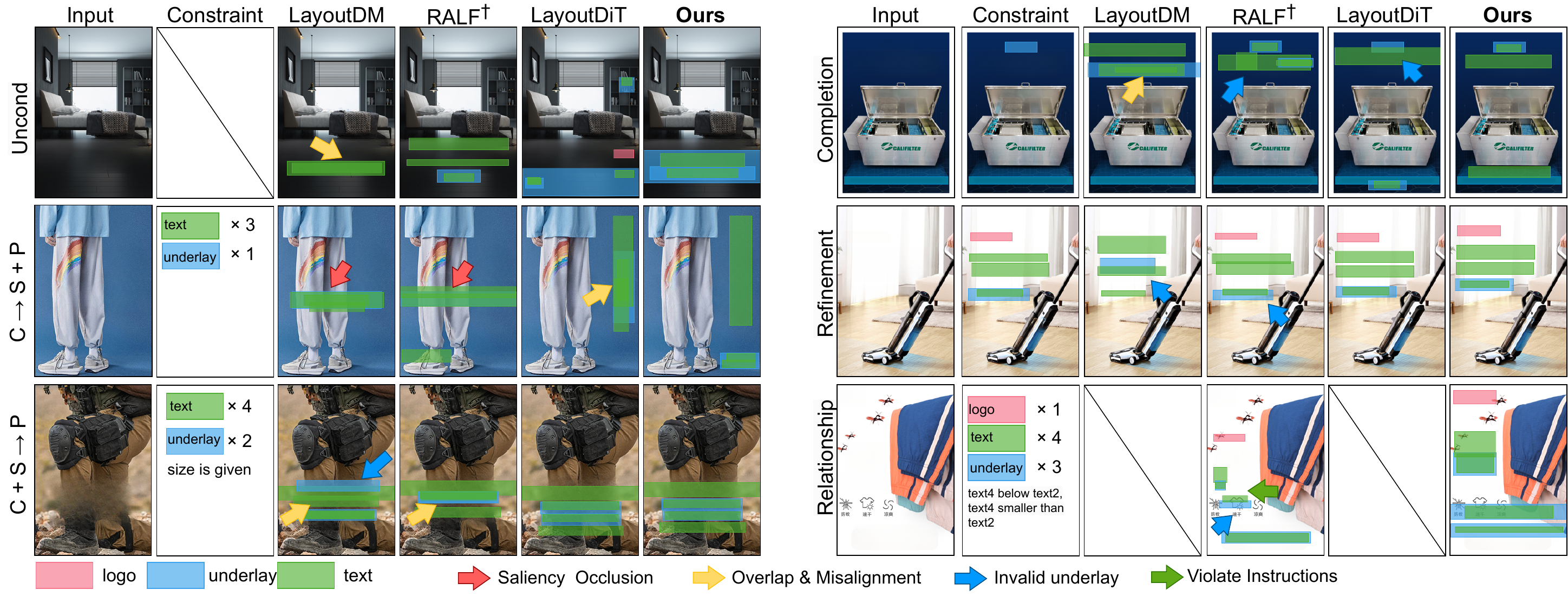}
    \vspace{-0.5cm}
    \caption{%
Visual comparison  with baselines on the PKU annotated test split.
    }
    \vspace{-0.5cm}
    \label{fig:quan}
\end{figure*}

\subsection{Ablation Study}

To validate our design choices, we conduct extensive ablation studies using the PKU annotated dataset.


\noindent \textbf{Strategy for integrating image and partial conditions.} 
To validate our unified training approach and the Dual-Path architecture, we compare several modality interaction schemes in Tab.~\ref{tab:unifiedtrainingmmdit}. The schemes are:

\begin{itemize}[leftmargin=1.5em]

\item [(a)] {Add}: Adding mask embeddings to layout embeddings.

\item [(b)]{Cross-Attn}: Using cross-attention for modal interaction.

\item [(c)]{Single-Path}: Using one MM-DiT branch to merge all three (image $\mathbf{F}_I$, layout $\mathbf{F}_L$, and mask $\mathbf{F}_M$) features.

\item [(d)]{Pre-Attn}: A variant of our model where visual ($\mathbf{F}_I$) and mask ($\mathbf{F}_M$) features first interact through an additional MM-Attention layer before their respective parallel interactions with the layout ($\mathbf{F}_L$) features; 

\item [(e)]{Dual-Path (Ours)}: The proposed approach using two parallel MM-DiT branches (one for $\mathbf{F}_I$-$\mathbf{F}_L$ interaction, one for $\mathbf{F}_M$-$\mathbf{F}_L$ interaction).
 
\item [(f)]{Task-specific}: Uses our proposed Dual-Path 
architecture (e), but is trained independently on each task, serving as a strong baseline.
\end{itemize}

\input{table/abl_uni}

\input{table/abl_rel_inj}

As shown in Tab.~\ref{tab:unifiedtrainingmmdit}, our unified \textit{Dual-Path} scheme achieves the best overall generation quality. The \textit{Single-Path} scheme underperforms, likely due to the optimization challenges of fusing three disparate modalities in a single attention mechanism~\cite{zhang2024creatilayout}. The \textit{Pre-Attn} and \textit{Cross-Attn} schemes also yield no improvement, suggesting that direct, explicit interaction between visual and mask modalities is not essential. The most critical relationships are those between the layout and the visual content, and between the layout and the constraint mask, which our \textit{Dual-Path} approach effectively captures in parallel.

Notably, our unified \textit{Dual-Path} model outperforms the \textit{Task-specific} baseline. This superiority stems from two factors: the effectiveness of our \textit{Dual-Path} design in handling diverse conditions, and the robust, generalized feature representation learned through unified training. This shared knowledge acts as a form of regularization, enabling it to surpass even those trained exclusively on a single task.


\noindent \textbf{Strategy for introducing relational constraints.} 
We ablate strategies for incorporating the relation task in Tab.~\ref{tab:relationshipinjection}, comparing three approaches:

\begin{itemize}[leftmargin=1.5em]
\item [(a)]{Base Model}: The model pre-trained on all tasks except the relation task, serving as our baseline.
\item [(b)]{Joint Training}: A model trained from scratch on all tasks, including the relation task, in a single stage. 
\item [(c)]{Full-param}: The Base Model fine-tuned for the relation task with full parameter updates. 
\item [(d)]{LoRA FT}: Our proposed method, where the Base Model is fine-tuned for the relation task using LoRA.
\end{itemize}

\begin{figure}[t]
\begin{center}
\includegraphics[width=\linewidth]{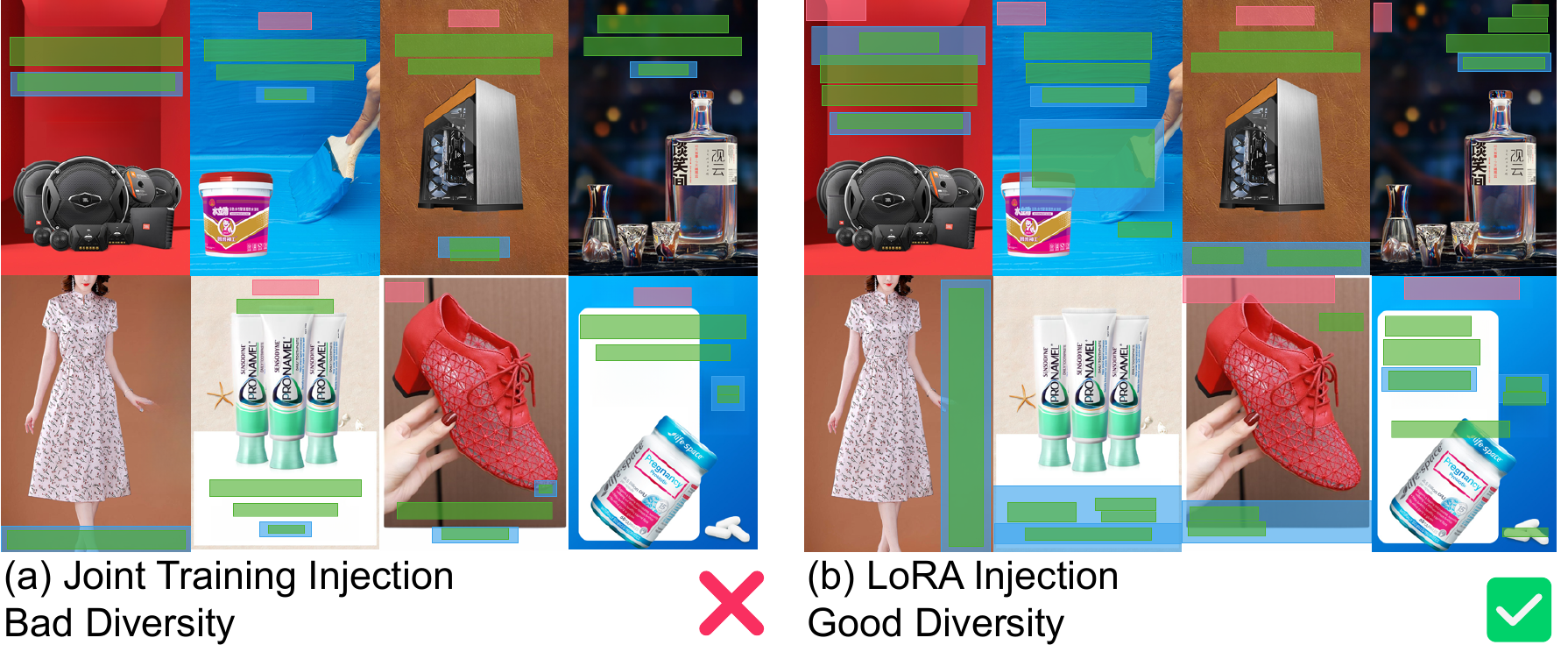}
\end{center}
\vspace{-0.5cm}

   \caption{Visual ablation study on the relationship injection strategy on the PKU dataset.}
\vspace{-0.5cm}

\label{fig:abltune} 
\end{figure}

We observed that the \textit{Joint Training} strategy significantly degrades quality on other tasks. As shown in Tab.~\ref{tab:relationshipinjection} and Fig.~\ref{fig:abltune}, this approach leads to poorer generation diversity (mode collapse), corresponding to a much higher FID. This suggests a conflict between the discrete, complex relation constraints and the generative priors for other tasks. By contrast, \textit{LoRA FT} achieves the best balance: it maintains strong relation adherence (Vio\% 22.4) significantly better than external baselines like RALF (34.2\%) while achieving the best performance on the unconditional task. This shows that \textit{LoRA FT} preserves and even improves upon the \textit{Base Model}'s generative capabilities.

We attribute this success to two factors. First, the two-stage (pre-train then fine-tune) approach avoids the optimization conflict between the discrete relation definitions and the continuous denoising process of other tasks. Second, the relation task narrows the valid solution space. LoRA fine-tuning is adept at making targeted adjustments to meet these new constraints without catastrophically disrupting the model's strong, pre-trained generative priors. 

\input{table/abl_loss}

\noindent \textbf{Effect of relational and content constraints.} 
Tab.~\ref{tab:constraints} presents the ablation results for the proposed $\mathcal{L}_{rel}$ and $\mathcal{L}_{ctn}$ losses. The results show that $\mathcal{L}_{rel}$ significantly reduces relational violations, while $\mathcal{L}_{ctn}$ effectively prevents collisions between layouts and high-saliency background regions. Further ablation studies on the loss parameters are provided in the supplementary material.




\subsection{Qualitative Results}

Fig.~\ref{fig:quan} compares UniLayDiff with baseline models on various tasks. UniLayDiff produces coherent, visually appealing layouts with minimal overlap and effective underlays, demonstrating strong content‑aware placement. It also faithfully enforces relation constraints, accurately modeling complex spatial and size relationships. Additional examples are provided in the supplementary material.

%% file: table/uncond_gen.tex
\begin{table*}
    [ht]
    \centering
    \setlength{\tabcolsep}{3pt} 
    \fontsize{10}{10}\selectfont
    \resizebox{0.9\linewidth}{!}{%
    \begin{tabular}{lccccccccccccc}
        \toprule
        & & \multicolumn{6}{c}{PKU annotated} & \multicolumn{6}{c}{CGL annotated} \\
        \cmidrule(lr){3-8} \cmidrule(lr){9-14}
        Method & \#Params & \multicolumn{2}{c}{Content} & \multicolumn{4}{c}{Graphic} & \multicolumn{2}{c}{Content} & \multicolumn{4}{c}{Graphic} \\
        \cmidrule(lr){3-4} \cmidrule(lr){5-8} \cmidrule(lr){9-10} \cmidrule(lr){11-14}
        & & Occ~$\downarrow$ & Rea~$\downarrow$ & Und\mysub{L}~$\uparrow$ & Und\mysub{S}~$\uparrow$ & Ove~$\downarrow$ & FID*~$\downarrow$ & Occ~$\downarrow$ & Rea~$\downarrow$ & Und\mysub{L}~$\uparrow$ & Und\mysub{S}~$\uparrow$ & Ove~$\downarrow$ & FID*~$\downarrow$ \\
        \midrule
        \textit{\small LLM-Based}& & & & & & & & & \\
        PosterLlama  & 7B & 0.149  & 0.0177 & 0.999 & \textbf{0.998} & 0.0008 & - & 0.1647 & 0.0184 & 0.998 & 0.989 & 0.0006 & - \\
        LayoutPrompter  & 8B & 0.227 & 0.0309 & 0.409 & 0.161 & 0.0017 & - & 0.2380 & 0.0327 & 0.383 & 0.174 & 0.0017 & - \\
        PosterO  & 8B & 0.135 & 0.0171 & 0.989 & 0.921 & 0.0006 & \textbf{2.71} & 0.137 & 0.0161 & 0.984 & 0.934 & 0.0006 & 5.17 \\
        \midrule
        \textit{\small  Non-LLM-based}& & & & & & & & & \\
         CGL-GAN & 41M & 0.176 & 0.0167 & 0.745 & 0.333 & 0.1071 & 37.53 & 0.196 & 0.0249 & 0.729 & 0.310 & 0.2574 & 68.13 \\
         DS-GAN & 30M & 0.159 & 0.0161 & 0.828 & 0.575 & 0.0353 & 12.45 & 0.153 & 0.0198 & 0.886 & 0.545 & 0.0415 & 45.27 \\
         ICVT & 50M & 0.280 & 0.0205 & 0.446 & 0.296 & 0.3155 & 43.18 & 0.227 & 0.0267 & 0.518 & 0.345 & 0.2337 & 67.44 \\
         LayoutDM$^{\dagger}$ & 43M & 0.151 & 0.0161 & 0.672 & 0.260 & 0.3224 & 24.98 & 0.152 & 0.0168 & 0.892 & 0.778 & 0.0325 & 2.62 \\
         RALF & 43M & 0.138 & 0.0126 & 0.977 & 0.907 & 0.0069 & 3.41 & 0.140 & 0.0141 & 0.994 & 0.980 & 0.0046 & \textbf{2.27} \\
         LayoutDiT  & 48M & \textbf{0.111} & 0.0125 & \textbf{0.999} & 0.996 & 0.0012 & 5.79 & \textbf{0.116} & 0.0151 & 0.998 & 0.989 & 0.0014 & 7.34 \\
         Scan-and-Print & 26M & 0.117  & 0.0128 & 0.973  & 0.963  & 0.0087  & 4.80 & 0.133 & \textbf{0.0126} & 0.970 & 0.963  & 0.0034 & 3.63 \\
        \rowcolor{lightblue} \textbf{UniLayDiff(Ours)} & 59M& 0.115 & \textbf{0.0114} & \textbf{0.999} & 0.996 & \textbf{0.0005} & 3.15 & 0.119 & 0.0127 & \textbf{0.999} & \textbf{0.993} & \textbf{0.0004} & 3.20 \\
        \bottomrule
    \end{tabular}
    }
\vspace{-0.5em}
    \captionof{table}{Comparison of unconditional generation results on PKU and CGL. * Indicates results we reproduced.} 
\vspace{-0.5em}
    \label{tab:res_uncond} 
\end{table*}

%% file: table/cond_gen.tex
\begin{table*}
\centering
\setlength{\tabcolsep}{3pt} 
\fontsize{10}{10}\selectfont
\resizebox{0.9\linewidth}{!}{
    \begin{tabular}{llcccccccccccc}
        \toprule
        \multirow{3}{*}{Task} & \multirow{3}{*}{Method} & \multicolumn{6}{c}{PKU annotated} & \multicolumn{6}{c}{CGL annotated} \\
        \cmidrule(lr){3-8} \cmidrule(lr){9-14}
        & & \multicolumn{2}{c}{Content} & \multicolumn{4}{c}{Graphic} & \multicolumn{2}{c}{Content} & \multicolumn{4}{c}{Graphic} \\
        \cmidrule(lr){3-4} \cmidrule(lr){5-8} \cmidrule(lr){9-10} \cmidrule(lr){11-14}
        & & Occ~$\downarrow$ & Rea~$\downarrow$ & Und\mysub{L}~$\uparrow$ & Und\mysub{S}~$\uparrow$ & Ove~$\downarrow$ & FID*~$\downarrow$ & Occ~$\downarrow$ & Rea~$\downarrow$ & Und\mysub{L}~$\uparrow$ & Und\mysub{S}~$\uparrow$ & Ove~$\downarrow$ & FID*~$\downarrow$ \\
        \midrule

        \multirow{4}{*}{\shortstack[l]{\textbf{C $\rightarrow$ S + P}}} 
        & LayoutDM$^{\dagger}$ & 0.161 & 0.0171 & 0.669 & 0.274 & 0.2949 & 22.19 & 0.141 & 0.0150 & 0.924 & 0.830 & 0.0332 & 4.70 \\
        & RALF & 0.136 & 0.0127 & 0.967 & 0.890 & 0.0091 & \textbf{2.36} & 0.140 & 0.0142 & 0.987 & 0.969 & 0.0056 & \textbf{0.62} \\
        & LayoutDiT & 0.121 & \textbf{0.0123} & \textbf{0.989} & 0.962 & 0.0025 & 3.31 & \textbf{0.117} & 0.0137 & 0.990 & 0.959 & 0.0077 & 3.84 \\
        & Scan-and-Print & \textbf{0.116} & 0.0131 & 0.965 & 0.952 & 0.0136 & 4.57 & 0.133 & 0.0133 & 0.963 & 0.956 & 0.0080 & 2.14    \\
        \rowcolor{lightblue} \cellcolor{white} & \textbf{UniLayDiff} & 0.122 & 0.0129 & \textbf{0.989} & \textbf{0.963} & \textbf{0.0018} & 3.09 & 0.122 & \textbf{0.0126} & \textbf{0.991} & \textbf{0.980} & \textbf{0.0015} & 1.72 \\
        \midrule

        \multirow{4}{*}{\shortstack[l]{\textbf{C + S $\rightarrow$ P}}} 
        & LayoutDM$^{\dagger}$ & 0.162 & 0.0152 & 0.755 & 0.563 & 0.2063 & 25.25 & 0.142 & 0.0149 & 0.922 & 0.834 & 0.0268 & 2.73 \\
        & RALF & 0.138 & 0.0123 & 0.945 & 0.876 & 0.0113 & 0.61 & 0.143 & 0.0148 & 0.984 & 0.957 & 0.0051 &     \textbf{0.24} \\
        & LayoutDiT & 0.133 & \textbf{0.0122} & 0.986 & \textbf{0.922} & 0.0059 & \textbf{0.59} & \textbf{0.127} & 0.0131 & \textbf{0.988} & 0.904 & 0.0075 & 1.20 \\
        \rowcolor{lightblue} \cellcolor{white}& \textbf{UniLayDiff} & \textbf{0.128} & 0.0132 & \textbf{0.987} & 0.918 & \textbf{0.0046} & 0.66 & 0.132 & \textbf{0.0127} & 0.986 & \textbf{0.945} & \textbf{0.0036} & 0.70 \\
        \midrule

        \multirow{4}{*}{\textbf{Completion}} 
        & LayoutDM$^{\dagger}$ & 0.141 & 0.0148 & 0.549 & 0.235 & 0.1822 & 22.62 & 0.142 & 0.0149 & 0.877 & 0.757 & 0.0247 & 3.45 \\
        & RALF & 0.137 & 0.0132 & 0.970 & 0.905 & 0.0139 & \textbf{1.46} & 0.141 & 0.0144 & \textbf{0.988} & 0.966 & 0.0042 & \textbf{0.95} \\
        & LayoutDiT & 0.119 & 0.0135 & 0.981 & 0.951 & 0.0027 & 3.80 & 0.123 & 0.0142 & 0.978 & 0.933 & 0.0039 & 4.76 \\
        \rowcolor{lightblue} \cellcolor{white}& \textbf{UniLayDiff} & \textbf{0.116} & \textbf{0.0128} & \textbf{0.985} & \textbf{0.974} & \textbf{0.0021} & 2.69 & \textbf{0.122} & \textbf{0.0136} & 0.987 & \textbf{0.971} & \textbf{0.0028} & 2.31 \\
        \midrule

        \multirow{4}{*}{\textbf{Refinement}} 
        & LayoutDM$^{\dagger}$ & 0.129 & 0.0112 & 0.910 & 0.599 & 0.0054 & 2.59 & 0.143 & 0.0145 & 0.890 & 0.614 & 0.0036 & 2.36 \\
        & RALF & 0.128 & 0.0103 & 0.989 & 0.951 & 0.0047 & \textbf{0.11} & 0.142 & 0.0136 & 0.994 & 0.980 & 0.0024 & \textbf{0.19} \\
        & LayoutDiT & 0.127 & \textbf{0.0094} & 0.990 & 0.965 & \textbf{0.0009} & 0.93 & 0.140 & \textbf{0.0109} & 0.987 & 0.946 & 0.0019 & 0.47 \\
        \rowcolor{lightblue} \cellcolor{white}& \textbf{UniLayDiff} & \textbf{0.123} & 0.0102 & \textbf{0.992} & \textbf{0.973} & 0.0010 & 0.76 & \textbf{0.136} & 0.0115 & \textbf{0.995} & \textbf{0.983} & \textbf{0.0013} & 0.51 \\
        \midrule

        \multirow{2}{*}{\textbf{Relationship*}} 
        & RALF & 0.124 & 0.0138 & 0.960 & 0.869 & 0.0085 & 2.41 & 0.126 & 0.0152 & 0.975 & 0.954 & 0.0060 & \textbf{0.73} \\
         & \textbf{UniLayDiff}\cellcolor{lightblue} & \textbf{0.119}\cellcolor{lightblue} & \textbf{0.0131}\cellcolor{lightblue} & \textbf{0.971}\cellcolor{lightblue} & \textbf{0.883}\cellcolor{lightblue} & \textbf{0.0049}\cellcolor{lightblue} & \textbf{2.15}\cellcolor{lightblue} & \textbf{0.121}\cellcolor{lightblue} & \textbf{0.0137}\cellcolor{lightblue} & \textbf{0.984}\cellcolor{lightblue} & \textbf{0.958}\cellcolor{lightblue} & \textbf{0.0043}\cellcolor{lightblue} & 1.74\cellcolor{lightblue} \\
        \bottomrule
    \end{tabular}
}
\vspace{-0.5em}
    \captionof{table}{Comparison of conditional generation results on PKU and CGL. * Indicates results we reproduced.} 
\vspace{-1.25\baselineskip}
    \label{tab:res_cond} 
\end{table*}

%% file: table/abl_uni.tex
\begin{table}[t]
\centering
\setlength{\tabcolsep}{3pt} 
\fontsize{10}{10}\selectfont
\resizebox{\linewidth}{!}{
\begin{tabular}{@{}lllccccc@{}}
\toprule
Tasks & \multicolumn{2}{l}{Method} & Occ~$\downarrow$ & Rea~$\downarrow$ & Und\mysub{S}~$\uparrow$ & Ove~$\downarrow$ & FID~$\downarrow$ \\ \midrule
\multirow{5}{*}{\textbf{C $\rightarrow$ S + P}} & \multicolumn{2}{l}{Task-specific} & 0.131 & 0.0129 & 0.943 & \textbf{0.0011} & 3.35\\
\cmidrule(l){2-8}
 & \multirow{4}{*}{Unified} & Add & 0.128 & \textbf{0.0127} & 0.947 & 0.0015 & 3.74\\
 & & Cross-Attn & 0.129 & 0.0134 & 0.957 & 0.0016 & 4.53\\
 & & Pre-Attn & 0.130 & 0.0132 & 0.951 & 0.0022 & 3.42  \\
 & & Single-Path & 0.132& 0.0132 & 0.949 & 0.0018 & 4.62\\
 \rowcolor{lightblue}\cellcolor{white}& \cellcolor{white}& Dual-Path & \textbf{0.126} & 0.0130 & \textbf{0.963} & 0.0013 & \textbf{3.11}\\ \midrule
\multirow{5}{*}{\textbf{C + S $\rightarrow$ P}} & \multicolumn{2}{l}{Task-specific} & 0.138 & 0.0135 & 0.896 & 0.0033 & \textbf{0.39} \\
\cmidrule(l){2-8}
 & \multirow{4}{*}{Unified} & Add & 0.134 & \textbf{0.0128} & 0.917 & 0.0037 & 0.54\\
 & & Cross-Attn & 0.132 & 0.0132 & 0.914 & 0.0035 & 0.53 \\
 & & Pre-Attn & 0.137 & 0.0134 & 0.904 & 0.0033 & 0.59 \\
 & & Single-Path & \textbf{0.129} & 0.0139 & 0.907 & 0.0042 & 0.63 \\
 \rowcolor{lightblue}\cellcolor{white}&\cellcolor{white} & Dual-Path & 0.135 & 0.0134 & \textbf{0.920} & \textbf{0.0029} & \textbf{0.47}\\ \midrule
\multirow{5}{*}{\textbf{Completion}} & \multicolumn{2}{l}{Task-specific} & 0.124 & \textbf{0.0127} & 0.968 & 0.0016 & 4.09 \\
\cmidrule(l){2-8}
 & \multirow{4}{*}{Unified} & Add & 0.118 &  0.0134  & 0.974 &  0.0014  & 4.18 \\
 & & Cross-Attn & 0.124 & 0.0139 & \textbf{0.975} & 0.0023 & 6.06 \\
 & & Pre-Attn & 0.135 & 0.0133 & 0.952 & 0.0031 & 4.02 \\
 & & Single-Path & 0.129 & 0.0132 & 0.946 & 0.0016 & 5.95 \\
 \rowcolor{lightblue}\cellcolor{white}&\cellcolor{white} & Dual-Path & \textbf{0.117} & \textbf{0.0127} & 0.971 & \textbf{0.0011} & \textbf{2.98}\\ \bottomrule
\end{tabular}
 }
\vspace{-0.3cm}
\caption{Ablation study on unified training with different architectures for partial constraint mask guidance. Relation injection is omitted as it is added during fine-tuning in our method.}
\vspace{-0.3cm}
\label{tab:unifiedtrainingmmdit}
\vspace{0.3cm}
\end{table}

%% file: table/abl_rel_inj.tex
\begin{table}[t]
\setlength{\tabcolsep}{3pt} 
\centering
\fontsize{10}{10}\selectfont
\resizebox{\linewidth}{!}{
\begin{tabular}{@{}c c c c c c c c @{} }
\toprule
Task & Relation inj. & Occ~$\downarrow$ & Rea~$\downarrow$ & Und\mysub{S}~$\uparrow$ & Ove~$\downarrow$ & FID~$\downarrow$ & Vio\%$\downarrow$ \\
\midrule
\multirow{3}{*}{\textbf{Uncond}} 
&  Base Model & 0.122 & 0.0120 & 0.992 & \textbf{0.0005} & 3.37  & --- \\
&  Joint Training & 0.111 & 0.0127 & 0.985 & 0.0013 & 9.71  & --- \\
~ & Full-param & \textbf{0.108} & 0.0125 & 0.994 & 0.0022 & 4.51 & --- \\
\rowcolor{lightblue}~\cellcolor{white} & LoRA FT & 0.115 & \textbf{0.0114} & \textbf{0.996} & \textbf{0.0005} & \textbf{3.15} & ---  \\
\cmidrule(lr){1-8}
\multirow{3}{*}{\textbf{Relationship}} & Joint Training & \textbf{0.112} & 0.0135 & 0.876 & 0.0054 & 6.18 &  \textbf{19.9}\\
~ & Full-param & 0.114 & 0.0136 & \textbf{0.904} & 0.0057 & 2.43 & 21.3 \\
\rowcolor{lightblue}~\cellcolor{white} & LoRA FT& 0.119 & \textbf{0.0131} & 0.883 & \textbf{0.0049} & \textbf{2.15} &  22.4\\
\bottomrule
\end{tabular}
}
\vspace{-0.2cm}
\caption{Ablation study on the relationship injection strategy on the PKU dataset. Unconditional generation serves as a metric for assessing performance retention on other tasks.}
\vspace{-0.3cm}
\label{tab:relationshipinjection}
\end{table}

%% file: table/abl_loss.tex
\begin{table}[htbp]
\centering
\fontsize{10}{10}\selectfont
\resizebox{\linewidth}{!}{
\begin{tabular}{cc cccccc}
\toprule
$\mathcal{L}_{\mathrm{rel}}$        & $\mathcal{L}_{\mathrm{ctn}}$        & Occ$\downarrow$ & Rea$\downarrow$ & Und$_S\uparrow$ & Ove~$\downarrow$ & FID~$\downarrow$ & Vio\%$\downarrow$ \\
\midrule
$\checkmark$ &              &          0.125       &         0.0129        &         0.904        &         0.0055         &        2.06          &         21.6        \\
             & $\checkmark$ &         0.118        &         0.0136        &         0.852        &         0.0043         &        2.18          &         38.7        \\
\midrule
\rowcolor{lightblue}$\checkmark$ & $\checkmark$ &        0.119         &        0.0131         &        0.883         &       0.0049           &          2.15        &         22.4        \\
\bottomrule
\end{tabular}
}
\vspace{-0.2cm}
\caption{Ablation study conducted during the fine-tuning stage to evaluate relational and content constraints on the relation subtask of the PKU dataset.}
\vspace{-0.5cm}
\label{tab:constraints}
\end{table}

%% file: sec/5_conclusion.tex
\section{Conclusion}
In this paper, we introduced UniLayDiff, a framework to successfully unify various content-aware layout generation tasks within a single, end-to-end trainable model. This unification is enabled by two core innovations: an MM-DiT based architecture that facilitates deep interaction among key modalities for unified content-aware generation, and a strategic fine-tuning process with additional constraints that employs LoRA fine-tuning to inject complex relational constraints without degrading the model's foundational generative capabilities. Extensive experiments demonstrate that UniLayDiff achieves state-of-the-art performance across various  tasks, moving automated graphic design toward more versatile, parameter‑efficient, and powerful solutions.

%% file: sec/X_suppl.tex
\clearpage
\setcounter{page}{1}
\maketitlesupplementary

\section*{A. Datasets}
In this section, we provide more details for PKU and CGL datasets.
\subsection*{A.1 Dataset Details}
The PKU dataset includes three element categories (logo, text, and underlay), while the CGL dataset also contains an additional ``embellishment'' category. In terms of scale, CGL is larger, with 60,548 annotated posters (each with both a layout and a corresponding image) and 1,000 unannotated canvases (images only). PKU contains 9,974 annotated posters and 905 unannotated canvases.

\subsection*{A.2 Data Splits}
Both the PKU and CGL datasets are divided into an annotated split and an unannotated split. To ensure a fair evaluation, we follow the data splitting protocol of RALF~\cite{horita2023retrieval} and LayoutDiT~\cite{li2024cgbdmcontentgraphicbalance}. The training, validation, and test sets are all partitioned from the annotated split; for the PKU dataset, this comprises 7734, 1000, and 1000 samples, while for the CGL dataset, it comprises 48544, 6002, and 6002 samples for training, validation, and testing, respectively. The unannotated split serves exclusively as an unannotated test set. Consequently, due to the lack of ground-truth layout annotations, this unannotated test set cannot be used for calculating metrics like FID and is only suitable for evaluating unconditional generation tasks.

\subsection*{A.3 Data Preprocessing}
The original PKU and CGL annotated datasets lack high-quality inpainted images and saliency maps. Therefore, subsequent works~\cite{li2024cgbdmcontentgraphicbalance, horita2023retrieval} commonly redo data preprocessing. The standard procedure involves using advanced inpainting algorithms to remove original graphic elements from the posters to create clean canvases~\cite{li2024cgbdmcontentgraphicbalance, horita2023retrieval}. For saliency detection, two methods are typically employed: ISNet~\cite{qin2022highly} and BASNet~\cite{qin2019basnet}. This results in two types of saliency maps: a more precise map that accurately captures object contours (from ISNet) and a slightly blurrier map that covers a broader salient area (from BASNet). The final saliency map $\mathbf{S}$ fed into the model is generated by taking the element-wise maximum of these two maps.

We would like to extend our gratitude to the authors of LayoutDiT~\cite{li2024cgbdmcontentgraphicbalance} for making their preprocessed data publicly available. In our work, we utilize their provided inpainted canvases, saliency maps, and saliency bounding boxes.

\section*{B. Implementation Details}

\subsection*{B.1 Architecture Details}
\noindent \textbf{MM-DiT.}
Our MM-DiT architecture consists of two parallel branches, each with a depth of 12. These branches are responsible for modeling layout-mask and layout-image interactions, respectively.
For the layout-mask MM-DiT, the input dimensions are $d_{\text{layout}} = 128$ and $d_{\text{mask}} = 128$. For the layout-image MM-DiT, the input dimensions are $d_{\text{image}} = 256$ and $d_{\text{layout}} = 128$. In each branch, the inputs are projected to a shared dimension of $d=512$ before the MM-Attention operation. The feed-forward network (FFN) dimension within each MM-DiT block is 1024.

\noindent \textbf{Image encoder.}
We use a ViT-based image encoder with an embedding dimension of 256, a depth of 12, and an FFN dimension of 2048. It takes an input image of size 384 $\times$ 256 and uses a patch size of 32.

\noindent \textbf{LoRA.}
We apply LoRA for fine-tuning, targeting only the layout-mask branch of our model. The adaptation is applied specifically to the linear projection matrices of the MM-Attention module within this branch, using a LoRA rank of \textbf{$r=4$} and a scaling factor of $\alpha=3$.




\subsection*{B.2 Constraints Details}
In this section, we present the details of the layout constraints, including the layout optimization loss and the weights of the losses.

\noindent \textbf{Layout optimization loss.}
To encourage the generation of layouts that adhere to established design paradigms, we incorporate a layout optimization loss, denoted as $\mathcal{L}_{\mathrm{lyt}}$. 
This objective aligns with the aesthetic constraints proposed in LACE~\cite{chen2024towards}, as both prioritize the rational internal arrangement of layout elements. 
Specifically, tailored for the unique underlay elements in content-aware layout generation, $\mathcal{L}_{\mathrm{lyt}}$ explicitly models the spatial constraints between underlays and other components. 
This encourages underlay elements to properly encompass their associated non-underlay counterparts.

For each underlay bounding box $\mathbf{b}_i$, we calculate its maximum coverage ratio, $C_{\max}(\mathbf{b}_i)$, over all non-underlay elements.

\begin{equation}
C_{\max}(\mathbf{b}_i) = \max_{j \in \mathcal{O}} \frac{\text{Area}(\mathbf{b}_i \cap \mathbf{b}_j)}{\text{Area}(\mathbf{b}_j)},
\end{equation}
where $C_{\max}(\mathbf{b}_i)$ is the maximum coverage ratio for a given underlay box $\mathbf{b}_i$; $\mathcal{O}$ is the set of all non-underlay elements; and $\mathbf{b}_j$ is an element from that set. All bounding boxes and their classifications (e.g., as underlay or non-underlay) are extracted from the predicted clean layout $\tilde{\mathbf{x}}_0$.

The final loss then penalizes incomplete coverage by averaging the coverage shortfall across all underlay elements.

\begin{equation}
\mathcal{L}_{\mathrm{lyt}} = \frac{1}{|\mathcal{B}|} \sum_{i \in \mathcal{B}} \left(1 - C_{\max}(\mathbf{b}_i)\right),
\end{equation}
where $\mathcal{L}_{\mathrm{lyt}}$ is the final loss value, and $\mathcal{B}$ is the set of all underlay elements, with $|\mathcal{B}|$ being its cardinality.

\noindent \textbf{Hyperparameters for additional constraints.}
For the total objective function $\mathcal{L}_{\mathrm{LoRA}}$ used in the LoRA fine-tuning stage, as formulated in Eq.~\ref{eq:loss_lora}, we configure the weighting coefficients as follows: $\lambda_{1}=0.2$ for the relational constraint term $\mathcal{L}_{\mathrm{rel}}$, $\lambda_{2}=0.4$ for the content constraint term $\mathcal{L}_{\mathrm{ctn}}$, and $\lambda_{3}=1$ for the layout optimization term $\mathcal{L}_{\mathrm{lyt}}$.
The internal hyperparameters for these loss functions are listed in Tab.~\ref{tab:loss_param}.
Moreover, the additional constraints are applied only during the first 30\% of the DDPM steps (the low-noise stage), while for the remaining 70\% of the steps, only $\mathcal{L}_{\mathrm{diff}}$ is applied.

\subsection*{B.3 Training Details}
We summarize the hyperparameters for both pre-training and fine-tuning UniLayDiff in Table~\ref{tab:train}. Moreover, the sampling proportions of different tasks for the pre-training and fine-tuning stages are shown in Table~\ref{tab:sample}. For optimization, we employ the Adam optimizer with parameters $\beta_1 = 0.9$ and $\beta_2 = 0.999$. The diffusion model (DDPM) is trained for a total of 1000 steps. For the final layout generation during inference, we use Denoising Diffusion Implicit Models (DDIM) sampling with 100 steps. In order to facilitate reproducibility, we fixed the random seed to 1 for all experiments (including both training and testing). All experiments were conducted on a single NVIDIA RTX 3090 GPU.

\input{table_supp/param}

\renewcommand{\thetable}{B}
\begin{table}[t]
    \centering
\resizebox{\linewidth}{!}{

\begin{tabular}{@{}c c c c c   @{}}
\toprule
Dataset & Stage &Learning-rate & Batch-size & Epoch 
\\
\midrule
\multirow{2}{*}{\makecell{CGL}} & 

Pre-training
& 1e-4 & 32 & 500
\\ 
~& LoRA Fine-tuning
& 2e-3 & 32 & 300
\\
\midrule
\multirow{2}{*}{\makecell{PKU}} &

Pre-training
& 2e-4 & 128 & 500
\\ 
~& LoRA Fine-tuning
& 3e-3 & 128 & 300
\\
\bottomrule
\end{tabular}
    }
    \caption{
Training hyperparameters for UniLayDiff on the PKU and CGL datasets.
    }
    \label{tab:train}
\vspace{-8pt}
  \end{table}

\renewcommand{\thetable}{C}
\begin{table}[t]
\centering
\setlength{\tabcolsep}{3pt}

\renewcommand{\arraystretch}{1.1} 
\resizebox{\linewidth}{!}{
\begin{tabular}{cccccc}
\toprule
Uncond & C $\rightarrow$ S + P & C + S $\rightarrow$ P & Completion & Relationship &  \\
\midrule

$2/5$  &   $1/5$   &   $1/5$   &   $1/5$   &   --        \\

$1/3$  &  $1/6$    &   $1/6$   &   $1/6$   &    $1/6$        \\

\bottomrule
\end{tabular}
}
\caption{Sampling proportions of different tasks during pre-training and fine-tuning.}
\label{tab:sample}
\vspace{-8pt}
\end{table}

  \input{table_supp/time}

\section*{C. Additional Experimental Results}

\input{table_supp/unanno}
\input{table_supp/relation}
\input{table_supp/new_task}

\subsection*{C.1  Computational Cost}

We compare the computational cost of UniLayDiff against several state-of-the-art (SOTA) models, including the autoregressive-based RALF~\cite{horita2023retrieval} and Scan-and-Print~\cite{Hsu-IJCAI2025-ScanandPrint}, the diffusion-based LayoutDiT~\cite{li2024cgbdmcontentgraphicbalance}, and the LLM-based PosterLlama~\cite{seol-2024-ECCV-posterllama} and PosterO~\cite{Hsu-CVPR2025-postero} in Tab~\ref{tab:time}. The comparison evaluates model parameters, inference time, and memory usage.

The parameter count and inference time of UniLayDiff are slightly higher than those of the diffusion-based LayoutDiT. However, UniLayDiff is designed as a unified model, offering superior parameter efficiency. Simultaneously, our model holds a significant advantage over LLM-based approaches regarding both inference speed and model size. Additionally, autoregressive models achieve faster inference in this context. This discrepancy arises because layout generation tasks typically involve a limited number of elements, resulting in fewer autoregressive iterations compared to the fixed sampling steps required by diffusion models.

\subsection*{C.2 Generalization and Robustness}
To validate the robustness of UniLayDiff, we conduct evaluations across three distinct dimensions: hybrid tasks, unannotated data, and cross-dataset scenarios.

\noindent \textbf{Generalization to hybrid tasks.}
We present the quantitative results on a set of hybrid generation tasks to further demonstrate the versatility and robustness of UniLayDiff. It is worth noting that previous task-specific models are incapable of handling these hybrid scenarios, as they are typically trained exclusively on single tasks.
In contrast, by flexibly combining the Partial Constraint Mask $\mathbf{M}_{P}$ and the relation matrix $\mathbf{M}_{R}$, our unified framework can seamlessly adapt to these complex scenarios.
Three hybrid tasks could be defined as follows:

\noindent \textit{Comp + (C $\rightarrow$ (S + P)).} 
This task combines layout completion with category-conditioned generation. 
Specifically, a subset of elements is fully specified (including category, size, and position), serving as the completion context. 
For the remaining elements, only their categories are provided.

\noindent \textit{Comp + ((C + S) $\rightarrow$ P).} 
This task integrates completion with category-and-size-conditioned generation. 
In addition to a set of fully defined elements (completion part), the categories and sizes of the remaining elements are given.

\noindent \textit{Comp + Relation.} 
This task imposes relational constraints on top of the standard completion task.


The quantitative results for these hybrid tasks are summarized in Tab.~\ref{tab:Hybrid}. 
These results highlight the versatility of UniLayDiff, demonstrating that our unified training strategy and MM-DiT architecture effectively empower the model to support a diverse range of constraint combinations.
\input{table_supp/cross}

\renewcommand{\thefigure}{A}
\begin{figure*}[t]
    \centering 
    \includegraphics[width=0.9\linewidth]{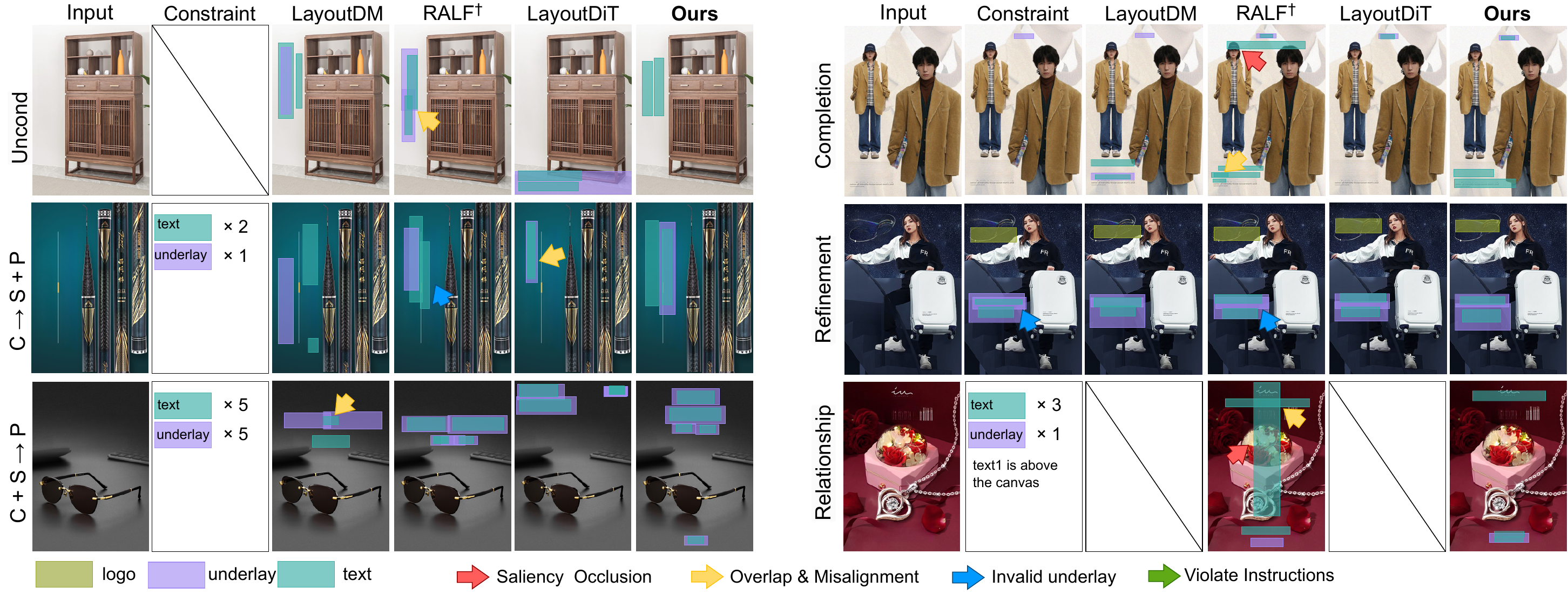}
    \vspace{-0.2cm}
    \caption{%
Visual comparison with baselines on the CGL annotated test split.
    }
    \vspace{-0.3cm}
    \label{fig:cgl0}
\end{figure*}

\noindent \textbf{Results on unannotated splits.}
Tab.~\ref{tab:unanno} reports unconditional generation performance on the unannotated test set. The results demonstrate that our method maintains strong performance on the challenging unannotated data.

\noindent \textbf{Cross-dataset Evaluation.}
The generalization of our method is verified through cross-dataset experiments. The model is trained on PKU and tested on CGL, and subsequently, trained on CGL and tested on PKU. As shown in Tab.~\ref{tab:cross_uncond}, UniLayDiff exceeded baselines on most metrics. These results confirm its robustness and adaptability to diverse real‑world images.

\renewcommand{\thetable}{I}
\begin{table}[t]
\centering
\setlength{\tabcolsep}{4pt}
\renewcommand{\arraystretch}{1.1} 
\resizebox{\linewidth}{!}{
\begin{tabular}{ccccccc}
\toprule
Scheme  & Occ~$\downarrow$ & Rea~$\downarrow$  & Und\mysub{S}~$\uparrow$ & Ove~$\downarrow$& FID~$\downarrow$ &Vio\%$\downarrow$\\
\midrule

All    &  \textbf{0.119}   &  0.0135  &   \textbf{0.885}   &  0.0053  &  2.43 & 25.1 \\

Layout-only    &  0.120  &    0.0135    &  0.876   &  \textbf{0.0047}  &   2.39  & 24.7 \\

Mask-only &   0.122   &   0.0133 &  0.874   &  0.0052   &  2.65   &  32.3\\

\rowcolor{lightblue} Diag(Ours)  &  \textbf{0.119}    &   \textbf{0.0131}   &  0.883    &   0.0049   &   \textbf{2.15}  & \textbf{22.4}\\

\bottomrule
\end{tabular}
}
\vspace{-4pt}
\caption{Ablation study on relative positional bias for relation injection using the PKU annotated dataset, evaluated using the relationship task.}
\label{tab:relbias}
\vspace{-8pt}
\end{table}

\input{table_supp/lora}

\subsection*{C.3 More Results for Relation Task}

Due to space constraints in the main text, we did not include the relation violation rate for the relation task in the main table. We report the full relation generation performance including the violation rate in Tab.~\ref{tab:relation}. Experimental results show that our unified model outperforms RALF in terms of relation violation rate as well as most of the other metrics, demonstrating the generative advantages of our model.

\subsection*{C.4 More Results on Ablation Study}
\noindent \textbf{Relative positional bias for relation injection.}
To validate the effectiveness of our masked relative positional bias strategy, we compare it with several alternative schemes in Tab.~\ref{tab:relbias}.
The schemes are:

\begin{itemize}[leftmargin=1.5em]

\item [(a)] {\textit{All}}: The relative positional bias is applied to all attention pairs across modalities.

\item [(b)] {\textit{Layout-only}}: The relative positional bias is restricted solely to the attention between layout elements.

\item [(c)] {\textit{Mask-only}}: The relative positional bias is restricted solely to the attention between mask elements.

\item [(d)] {\textit{Diag} }: Our proposed masked relative positional bias, which effectively models the relationships within each modality as shown in Eq.~\ref{eq:attnmask}.

\end{itemize}

As shown in Tab.~\ref{tab:relbias}, the \textit{Diag} scheme achieves the best performance, indicating its strong capability in relationship modeling.
Compared to the \textit{All} scheme, our \textit{Diag} approach is based on a clearer premise, as the concept of relative position between elements of different modalities is not always meaningful.


    

    

\noindent \textbf{LoRA fine-tuning strategy for relation.}
In UnilayDiff, LoRA fine-tuning is specifically applied to the Partial Constraint Mask branch ($\mathbf{F}_M$-$\mathbf{F}_L$) rather than both branches of the MM-DiT (Sec.~\ref{sec:LoRA}). To validate this design, we compare two different LoRA application strategies:
\begin{itemize}[leftmargin=1.5em]

    \item [(a)] \textit{Dual Branch FT}: LoRA fine-tuning is applied to both the Image-Layout ($\mathbf{F}_I$-$\mathbf{F}_L$) and Mask-Layout ($\mathbf{F}_M$-$\mathbf{F}_L$) branches.
    \item [(b)] \textit{Mask-Layout FT} (Ours): LoRA is applied exclusively to the Mask-Layout interaction branch ($\mathbf{F}_M$-$\mathbf{F}_L$).
\end{itemize}
The results are summarized in Tab.~\ref{tab:lora}. 
We observe that the \textit{Dual} strategy does not yield performance gains for the relationship task (Vio\%) compared to our \textit{Mask-only} strategy. 
Although \textit{Dual} shows a slight improvement in the Overlay metric (Ove), it leads to performance degradation in other metrics. 
This suggests that relational constraints are inherently structural and are most effectively injected through the mask branch.

\subsection*{C.5 Visual Comparison}
We provide additional visual examples of UniLayDiff in Fig.~\ref{fig:cgl0}, Fig.~\ref{fig:pku1}, Fig.~\ref{fig:cgl1}, and Fig.~\ref{fig:unanno}.
Among these, Fig.~\ref{fig:cgl0} shows a set of results generated by the model for each subtask on the CGL annotated split. In Fig.~\ref{fig:pku1} and Fig.~\ref{fig:cgl1}, we provide additional generated results on the PKU and CGL annotated splits to supplement the visualizations in Fig.~\ref{fig:cgl0} and the main text. Furthermore, we demonstrate the generation performance of UniLayDiff on real-world data (the unannotated split) in Fig.~\ref{fig:unanno}. The visual results show that UniLayDiff can generate high-quality and diverse layouts.

\clearpage


\renewcommand{\thefigure}{B}
\begin{figure*}[ht]
    \centering 
    \includegraphics[width=0.8\linewidth]{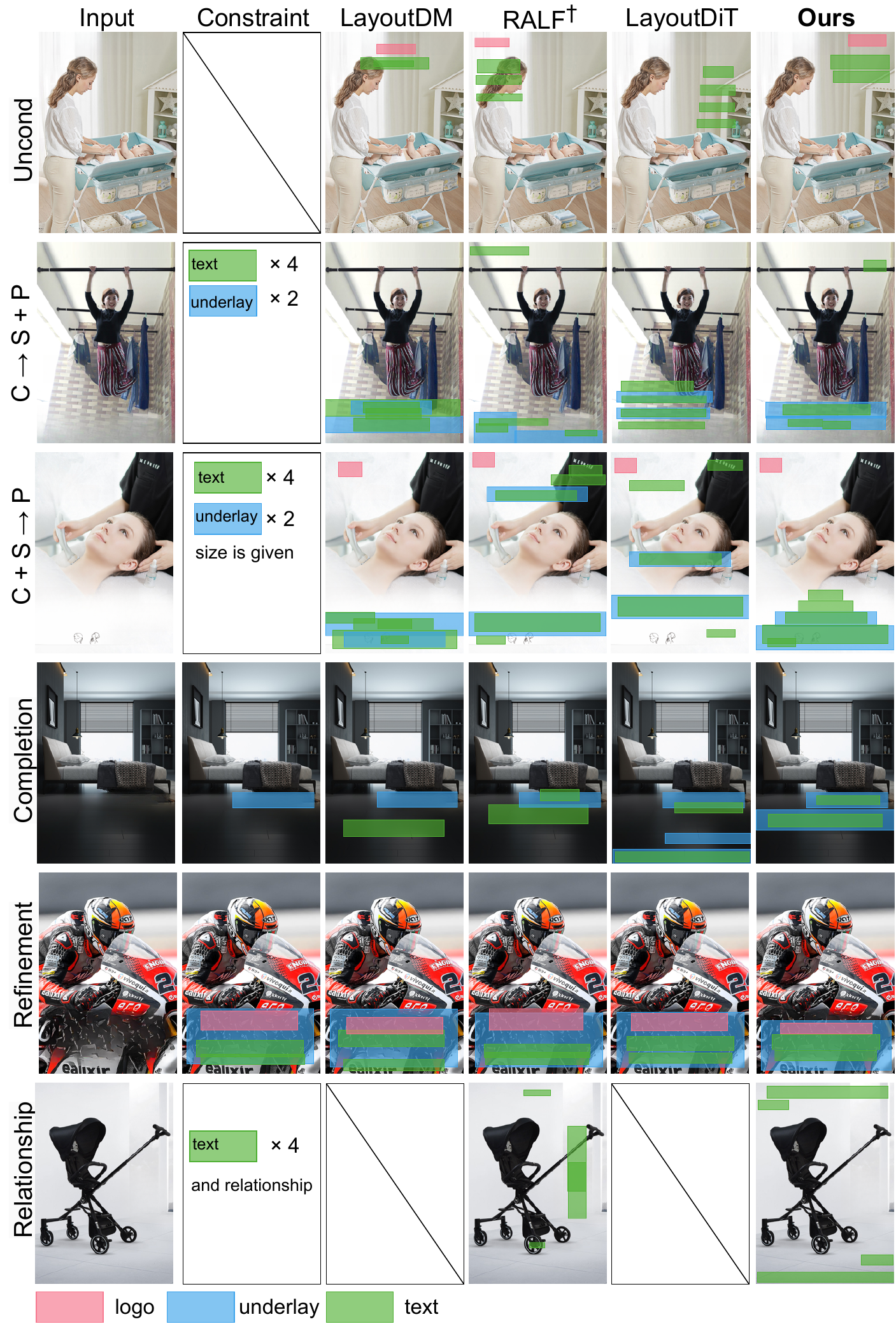}

    \caption{%
Visual comparison with baselines on the PKU annotated test split.
    }

    \label{fig:pku1}
\end{figure*}

\renewcommand{\thefigure}{C}
\begin{figure*}[t]
    \centering 
    \includegraphics[width=0.8\linewidth]{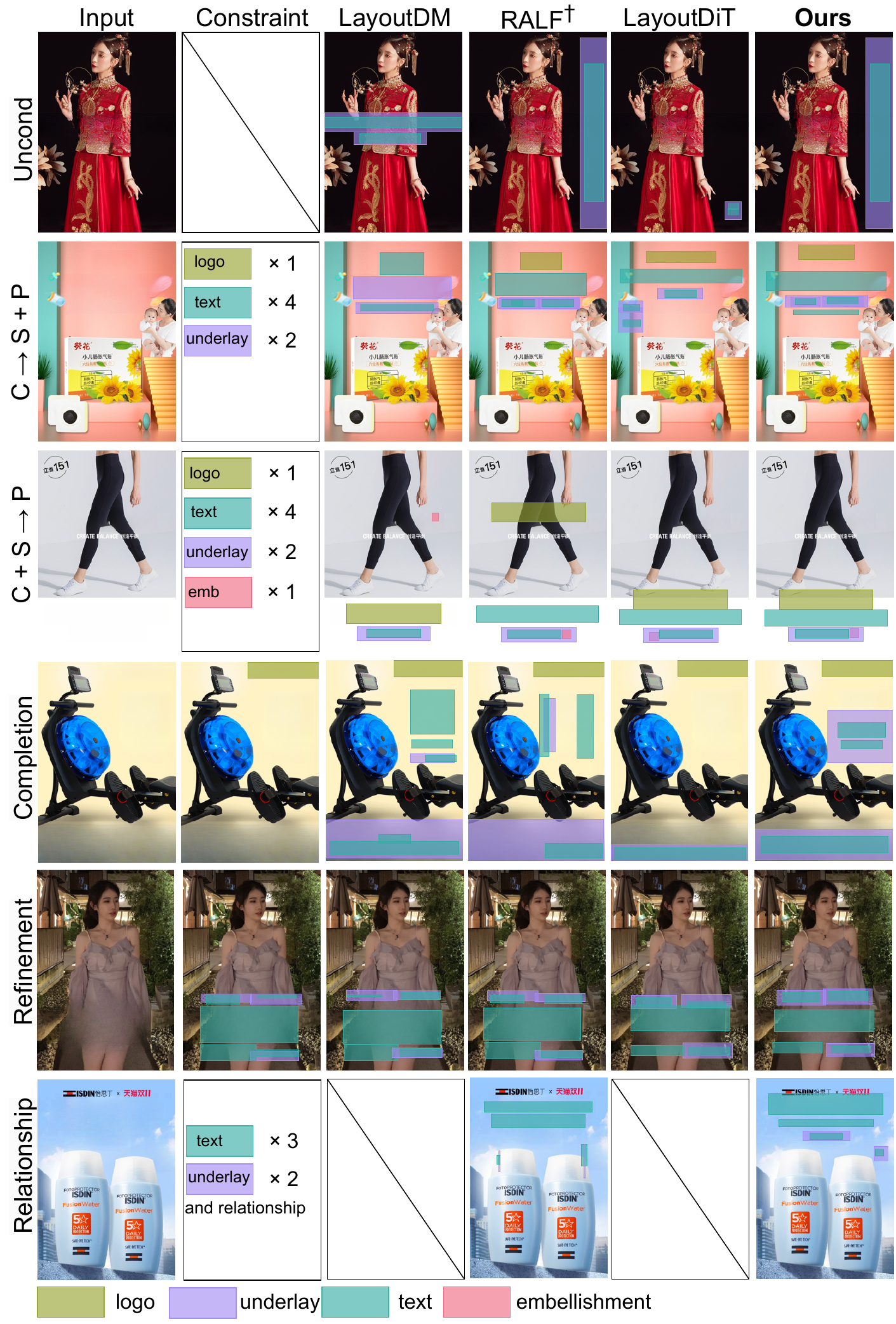}
    \caption{%
More visual comparison with baselines on the CGL annotated test split.
    }
    \label{fig:cgl1}
\end{figure*}

\renewcommand{\thefigure}{D}
\begin{figure*}[t]
    \centering 
    \includegraphics[width=0.7\linewidth]{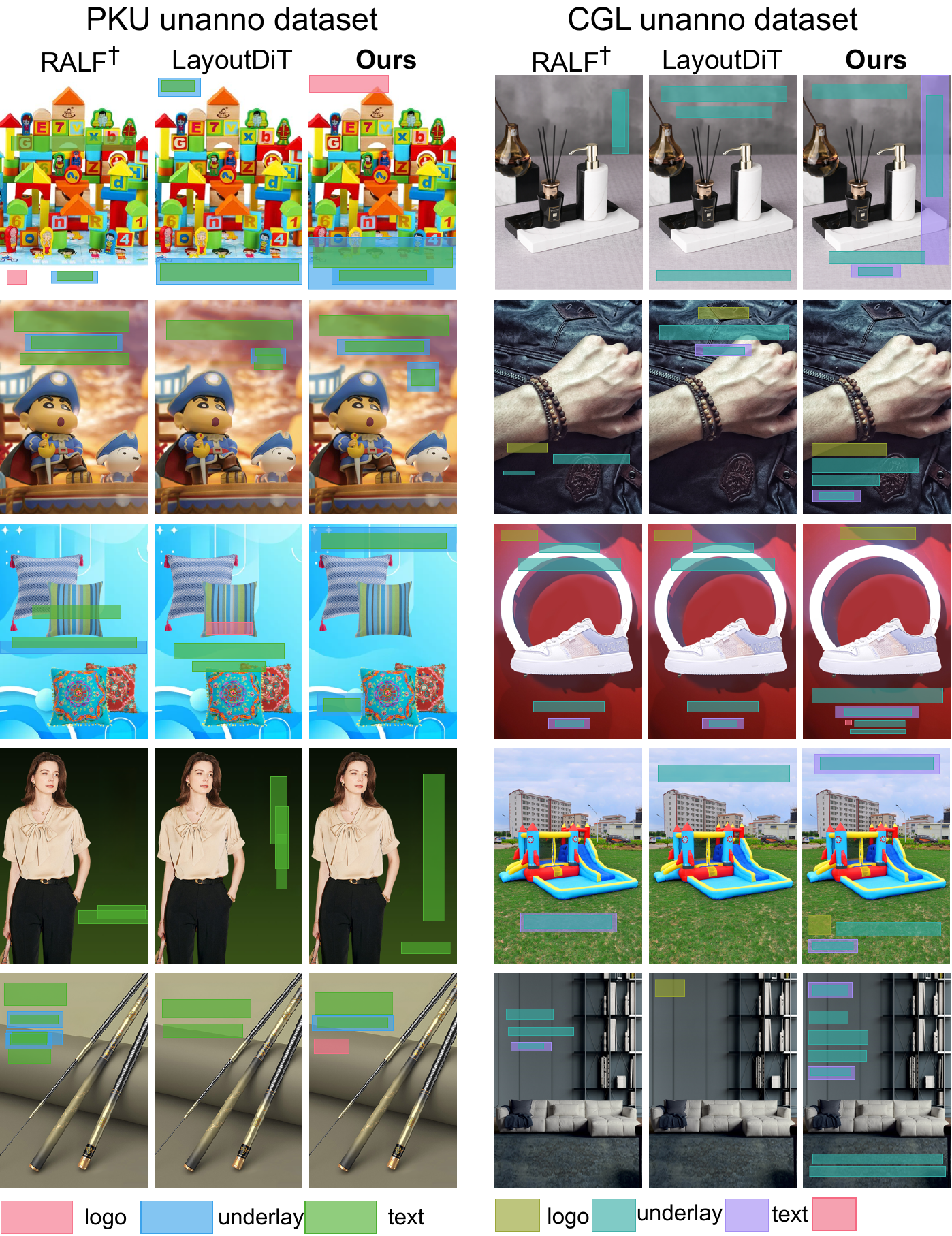}
    \caption{%
Visual comparison with baselines on PKU and CGL unannotated test splits.
    }
    \label{fig:unanno}
\end{figure*}


%% file: table_supp/param.tex
\renewcommand{\thetable}{A} 
\begin{table}[t]
\centering
\setlength{\tabcolsep}{10pt}

\renewcommand{\arraystretch}{1.2} 
\resizebox{0.8\linewidth}{!}{
\begin{tabular}{lcc}
\toprule
Hyperparameter & Symbol & Value \\
\midrule
Relational Tolerance Margin & $\alpha$ & 0.1 \\
Relational Temperature & $\tau_{\mathrm{rel}}$ & 0.6  \\
Sigmoid Temperature & $\tau_{\mathrm{sig}}$ & 1.0 \\
Stability Constant & $\delta_{\mathrm{eps}}$ & 1e-6 \\
\bottomrule
\end{tabular}
}
\caption{Hyperparameters included in the loss functions.}
\label{tab:loss_param}
\vspace{-8pt}
\end{table}

%% file: table_supp/time.tex
\renewcommand{\thetable}{D}
\begin{table}[t]
\centering
\setlength{\tabcolsep}{5pt}

\renewcommand{\arraystretch}{1.1} 
\resizebox{\linewidth}{!}{

\begin{tabular}{lccc} 
\toprule

& \#Param & Inference Memory & Inference Time \\ 
\midrule
\textit{\small  Autoregressive}& & & \\
RALF & 43M & 8.4GB & 15ms/sample \\
Scan-and-Print & 26M & 1.4GB & 5.8ms/sample \\
\midrule
\textit{\small  Diffusion-based}& & & \\
LayoutDiT & 48M & 1.3GB & 57ms/sample \\
UniLayDiff & 59M & 1.7GB & 93ms/sample \\
\midrule
\textit{\small  LLM-based}& & & \\
PosterLlama & 7B & 39GB & 6380ms/sample \\
PosterO & 8B & 23GB & 3881ms/sample \\
\bottomrule
\end{tabular}
}
\caption{Comparison of computational cost.}
\label{tab:time}
\vspace{-8pt}
\end{table}

%% file: table_supp/unanno.tex
\renewcommand{\thetable}{E}
\begin{table*}
    [ht]
    \centering
    \setlength{\tabcolsep}{5pt} 
    \fontsize{10}{10}\selectfont
    \resizebox{0.9\linewidth}{!}{%
    \begin{tabular}{lccccccccccc}
        \toprule
        & & \multicolumn{5}{c}{PKU annotated} & \multicolumn{5}{c}{CGL annotated} \\
        \cmidrule(lr){3-7} \cmidrule(lr){8-12}
        Method & \#Params & \multicolumn{2}{c}{Content} & \multicolumn{3}{c}{Graphic} & \multicolumn{2}{c}{Content} & \multicolumn{3}{c}{Graphic} \\
        \cmidrule(lr){3-4} \cmidrule(lr){5-7} \cmidrule(lr){8-9} \cmidrule(lr){10-12}
        & & Occ~$\downarrow$ & Rea~$\downarrow$ & Und\mysub{L}~$\uparrow$ & Und\mysub{S}~$\uparrow$ & Ove~$\downarrow$ & Occ~$\downarrow$ & Rea~$\downarrow$ & Und\mysub{L}~$\uparrow$ & Und\mysub{S}~$\uparrow$ & Ove~$\downarrow$ \\
        \midrule
        \textit{\small LLM-Based}& & & & & & & & & \\
        PosterLlama  & 7B & 0.163 & 0.0285 & 0.998 & 0.992 & 0.0006 & 0.468 & 0.0555 & 0.997 & 0.977 & 0.0014 \\
        LayoutPrompter  & 8B & 0.254 & 0.0412 & 0.405 & 0.162 & 0.0010 & 0.457 & 0.0644 & 0.323 & 0.114 & 0.0026 \\
        PosterO  & 8B & 0.144 & 0.0248 & 0.985 & 0.924 & \textbf{0.0004} & 0.291 & 0.0365 & 0.981 & 0.903 & \textbf{0.0004} \\
        \midrule
        \textit{\small  Non-LLM-based}& & & & & & & & & \\
         CGL-GAN & 41M & 0.213 & 0.0256 & 0.718 & 0.299 & 0.1034 & 0.496 & 0.0604 & 0.652 & 0.129 & 0.2478 \\
         DS-GAN & 30M & 0.179 & 0.0228 & 0.764 & 0.464 & 0.0368 & 0.407 & 0.0533 & 0.804 & 0.349 & 0.0831 \\
         ICVT & 50M & 0.295 & 0.0254 & 0.420 & 0.301 & 0.3167 & 0.464 & 0.0502 & 0.466 & 0.306 & 0.1959 \\
         LayoutDM$^{\dagger}$ & 43M & 0.153 & 0.0220 & 0.645 & 0.235 & 0.3051 & 0.432 & 0.0499 & 0.730 & 0.539 & 0.0979 \\
         RALF & 43M & 0.142 & 0.0190 & 0.965 & 0.878 & 0.0106 & \textbf{0.258} & \textbf{0.0319} & 0.986 & 0.935 & 0.0572 \\
         LayoutDiT  & 48M & 0.124 & 0.0161 & \textbf{0.999} & \textbf{0.994} & \textbf{0.0004} & 0.339 & 0.0358 & 0.998 & 0.989 & 0.0023 \\
         Scan-and-Print & 26M & \textbf{0.116} & 0.0181 & 0.983 & 0.970 & 0.0090 & 0.336 & 0.0323 & 0.985 & 0.974 & 0.0157 \\
        \rowcolor{lightblue} \textbf{UniLayDiff(Ours)} & 59M& 0.118 & \textbf{0.0155} & \textbf{0.999} & \textbf{0.994} & 0.0009 & 0.358 & 0.0335 & \textbf{0.999} & \textbf{0.994} & 0.0008 \\
        \bottomrule
    \end{tabular}
    }
\vspace{-0.5em}
    \captionof{table}{Unconstrained generation results on the PKU and CGL unannotated test split.} 
\vspace{-0.5em}
    \label{tab:unanno} 
\end{table*}

%% file: table_supp/relation.tex
\renewcommand{\thetable}{F}
\begin{table*}[!t]
\centering
\setlength{\tabcolsep}{3pt}
\resizebox{0.9\linewidth}{!}{
\begin{tabular}{lcccccccccccccc}
\toprule
\multirow{2}{*}{Method } & 
\multicolumn{7}{c}{PKU annotated} & \multicolumn{7}{c}{CGL annotated} \\
\cmidrule(lr){2-8} \cmidrule(lr){9-15}
& Occ$\downarrow$ & Rea$\downarrow$ 
& Und\mysub{L}$\uparrow$ & Und\mysub{S}$\uparrow$ & Ove$\downarrow$ & FID~$\downarrow$ &Vio\%$\downarrow$
& Occ$\downarrow$ & Rea$\downarrow$ 
& Und\mysub{L}$\uparrow$ & Und\mysub{S}$\uparrow$ & Ove$\downarrow$ & FID~$\downarrow$ &Vio\%$\downarrow$\\
\midrule
RALF & 0.124 & 0.0138 & 0.960 & 0.869 & 0.0085 & 2.41 & 34.2 & 0.126 & 0.0152 & 0.975 & 0.954 & 0.0060 & \textbf{0.73} & 36.1\\
\rowcolor{lightblue}\textbf{UniLayDiff(Ours)} & \textbf{0.119} & \textbf{0.0131} & \textbf{0.971} & \textbf{0.883} & \textbf{0.0049} & \textbf{2.15} & \textbf{22.4}& \textbf{0.121} & \textbf{0.0137} & \textbf{0.984} & \textbf{0.958} & \textbf{0.0043} & 1.74 & \textbf{21.6}\\
\bottomrule
\end{tabular}
}
\vspace{-6pt}
\caption{Generation results for relation task on the PKU and CGL annotated test split.}
\vspace{-10pt}
\label{tab:relation}
\end{table*}

%% file: table_supp/new_task.tex
\renewcommand{\thetable}{G}
\begin{table}
\centering
\setlength{\tabcolsep}{2pt} 
\fontsize{10}{10}\selectfont
\resizebox{1.0\linewidth}{!}{
    \renewcommand{\arraystretch}{1.2} 
    \begin{tabular}{lcccccc}
        \toprule

        \multirow{2}{*}{Hybrid Tasks} & \multicolumn{2}{c}{Content} & \multicolumn{4}{c}{Graphic} \\

        \cmidrule(lr){2-3} \cmidrule(lr){4-7}

        & Occ~$\downarrow$ & Rea~$\downarrow$ & Und\mysub{L}~$\uparrow$ & Und\mysub{S}~$\uparrow$ & Ove~$\downarrow$ & FID~$\downarrow$ \\
        \midrule

        Comp + (C $\rightarrow$ (S + P)) & 0.121 & 0.0130  & 0.981 & 0.951 & 0.003 & 2.34  \\
        Comp + ((C + S) $\rightarrow$ P) & 0.125 &  0.0132 & 0.974 & 0.894 & 0.005 & 0.76 \\
        Comp + Relation & 0.122 & 0.0129 & 0.974 & 0.894 & 0.0060 & 2.04 \\
        
        \bottomrule
    \end{tabular}
}
\vspace{-0.5em}

    \captionof{table}{Quantitative evaluation for the hybrid tasks on the PKU annotated test split.} 
\vspace{-1.25\baselineskip}
    \label{tab:Hybrid} 
\end{table}


        

%% file: table_supp/cross.tex
\renewcommand{\thetable}{H}
\begin{table}[t]
\centering
\setlength{\tabcolsep}{2pt}
\resizebox{\linewidth}{!}{
\begin{tabular}{@{}c c c c c c c c c c c c @{}}
\toprule
Train & Test & Method & Occ~$\downarrow$ & Rea~$\downarrow$ & Und\mysub{L}~$\uparrow$ & Und\mysub{S}~$\uparrow$ & Ove~$\downarrow$
\\
\midrule
\multirow{4}{*}{CGL} & \multirow{2}{*}{PKU-unanno} & RALF & 0.143&0.0170&0.989&0.958&0.0347  
\\ 
~ & ~ & LayoutDiT
& 0.115&0.0178&\textbf{0.998}&0.980& 0.0025
\\
\rowcolor{lightblue}~\cellcolor{white} & ~\cellcolor{white} & Ours
& \textbf{0.107} & \textbf{0.0165} & \textbf{0.998} & \textbf{0.986} & \textbf{0.0012}
\\
\cmidrule{2-8}
~ & \multirow{2}{*}{PKU-anno} &RALF&0.140& \textbf{0.0113}& 0.995&0.976&0.0078 
\\ 
~ & ~ & LayoutDiT
& 0.108&0.0142&0.995&0.987&0.0015
\\

\rowcolor{lightblue}~\cellcolor{white} & ~\cellcolor{white} & Ours
& \textbf{0.105} & \textbf{0.0113} & 0.996 & \textbf{0.990} & \textbf{0.0009}
\\

\midrule
\multirow{4}{*}{PKU} &
\multirow{2}{*}{CGL-unanno} &
RALF &0.342&0.0425&0.927&0.773& 0.0309  
\\ 
~&~& LayoutDiT
& \textbf{0.328}&0.0331&\textbf{1.000}&0.992 &0.0053
\\
\rowcolor{lightblue}~\cellcolor{white} & ~\cellcolor{white} & Ours
& 0.343 & \textbf{0.0275} & 0.999 & \textbf{0.993} & \textbf{0.0044}
\\
\cmidrule{2-8}
~ & \multirow{2}{*}{CGL-anno} &RALF&0.150&0.0170&0.974&0.902&0.0065
\\ 
~ & ~ & LayoutDiT
& 0.128&0.0147&\textbf{0.999}&\textbf{0.995}&
0.0040
\\

\rowcolor{lightblue}~\cellcolor{white} & ~\cellcolor{white} & Ours
& \textbf{0.123} & \textbf{0.0138} & \textbf{0.999} & 0.992 &  \textbf{0.0035}
\\

\bottomrule
\end{tabular}
}
\vspace{-5pt}
\captionof{table}{Cross-dataset evaluation setup: We train a model on PKU and test it on CGL, or vice versa.}
\label{tab:cross_uncond}
\vspace{-8pt}
\end{table}

%% file: table_supp/lora.tex
\renewcommand{\thetable}{J}
\begin{table}[t]
\centering
\setlength{\tabcolsep}{3pt}
\renewcommand{\arraystretch}{1.1} 
\resizebox{\linewidth}{!}{
\begin{tabular}{ccccccc}
\toprule
Scheme & Occ~$\downarrow$ & Rea~$\downarrow$ & Und\mysub{S}~$\uparrow$ & Ove~$\downarrow$& FID~$\downarrow$ &Vio\%~$\downarrow$\\
\midrule

Dual Branch FT    & \textbf{0.115} & \textbf{0.0128} & 0.879  & 0.0053  & 2.44 & 23.5 \\

\rowcolor{lightblue} Mask-Layout FT  & 0.119 & 0.0131 & \textbf{0.883} & \textbf{0.0049} & \textbf{2.15} & \textbf{22.4}\\

\bottomrule
\end{tabular}
}
\vspace{-4pt}
\caption{Ablation study on LoRA fine-tuning strategy for relation using the PKU annotated dataset.}
\label{tab:lora}
\vspace{-8pt}
\end{table}